\documentclass[review]{elsarticle}
\graphicspath{ {./figures/} }
\usepackage{hyperref}
\usepackage{float}
\usepackage{verbatim} 
\usepackage{apalike}
\restylefloat{figure}
\floatstyle{plaintop} 
\restylefloat{table}

\usepackage{amsmath}
\usepackage{amssymb}
\usepackage{subfig}
\usepackage{algorithm}
\usepackage{algorithmic}
\usepackage{booktabs}
\usepackage{xcolor}
\usepackage{float}
\usepackage{graphicx}
\usepackage{multirow}
\usepackage{hyperref}
\newtheorem{theorem}{Theorem}

\journal{Expert Systems with Applications}

\biboptions{authoryear}

\begin{document}
\begin{frontmatter}


\begin{titlepage}
\begin{center}
\vspace*{1cm}

\textbf{ \large Robustifying Diffusion-Denoised Smoothing Against Covariate Shift}

\vspace{1.5cm}

Ali Hedayatnia$^{a}$ (alihedayatnia@ut.ac.ir), Mostafa Tavassolipour$^a$ (tavassolipour@ut.ac.ir), Babak Nadjar Araabi$^a$ (araabi@ut.ac.ir), Abdol-Hossein Vahabie$^a$ (h.vahabie@ut.ac.ir) \\

\hspace{10pt}

\begin{flushleft}
\small  
$^a$ Department of Electrical and Computer Engineering, University of Tehran, Tehran, Iran 

\vspace{1cm}
\textbf{Corresponding author at: Department of Electrical and Computer Engineering, University of Tehran, Tehran, Iran.} \\
Mostafa Tavassolipour \\
Department of Electrical and Computer Engineering, University of Tehran, Tehran, Iran. \\
Tel: (+98) 2161115073 \\
Email: tavassolipour@ut.ac.ir \\
Abdol-Hossein Vahabie \\
Department of Electrical and Computer Engineering, University of Tehran, Tehran, Iran. \\
Tel: (+98) 9194211374 \\
Email: h.vahabie@ut.ac.ir

\end{flushleft}        
\end{center}
\end{titlepage}

\title{Robustifying Diffusion-Denoised Smoothing Against Covariate Shift}


\author[ECE_UT]{Ali Hedayatnia}
\ead{alihedayatnia@ut.ac.ir}

\author[ECE_UT]{Mostafa Tavassolipour\corref{cor1}}
\ead{tavassolipour@ut.ac.ir}

\author[ECE_UT]{Babak Nadjar Araabi}
\ead{araabi@ut.ac.ir}

\author[ECE_UT]{Abdol-Hossein Vahabie\corref{cor1}}
\ead{h.vahabie@ut.ac.ir}

\cortext[cor1]{Corresponding author.}
\address[ECE_UT]{Department of Electrical and Computer Engineering, University of Tehran, Tehran, Iran}


\begin{abstract}
Randomized smoothing is a well-established method for achieving certified robustness against $l_2$-adversarial perturbations. By incorporating a denoiser before the base classifier, pretrained classifiers can be seamlessly integrated into randomized smoothing without significant performance degradation. Among existing methods, Diffusion Denoised Smoothing—where a pretrained denoising diffusion model serves as the denoiser—has produced state-of-the-art results. However, we show that employing a denoising diffusion model introduces a covariate shift via misestimation of the added noise, ultimately degrading the smoothed classifier’s performance. To address this issue, we propose a novel adversarial objective function focused on the added noise of the denoising diffusion model. This approach is inspired by our understanding of the origin of the covariate shift. Our goal is to train the base classifier to ensure it is robust against the covariate shift introduced by the denoiser. Our method significantly improves certified accuracy across three standard classification benchmarks—MNIST, CIFAR-10, and ImageNet—achieving new state-of-the-art performance in $l_2$-adversarial perturbations. Our implementation is publicly available at \href{https://github.com/ahedayat/Robustifying-DDS-Against-Covariate-Shift}{this GitHub repository}
\end{abstract}


\end{frontmatter}

\section{Introduction}
Deep neural networks have achieved remarkable success across various tasks, including image classification \cite{yu2022coca, wortsman2022model, chen2022pali}, object detection \cite{zong2023detrs, wang2023internimage, su2023towards, oksuz2023mocae}, and semantic segmentation \cite{wang2023one, wang2023internimage, su2023towards, wang2023image}, often surpassing human performance. However, these models remain vulnerable to data that deviates from training data distribution. For instance, adding even imperceptible perturbations to a data can cause a well-trained classifier to misclassify it. Such perturbed inputs, known as adversarial examples \cite{szegedy2013intriguing, biggio2013evasion}, pose significant challenges for deploying neural networks in safety-critical applications.

The sensitivity of neural networks to input perturbations has made robustness a critical concern, leading to the development of various defense strategies. Among the most effective approaches is adversarial training \cite{mkadry2017towards, goodfellow2014explaining, huang2015learning, ding2018mma}, which generates adversarial examples during training and incorporates them into the learning process. Another interesting approach involves exposing classifiers to specific attacks during training. While such methods can improve the model's robustness to specific perturbation, they often lack generalizability \cite{tramer2019adversarial, kaufmann2019testing, schmidt2018adversarially, rice2020overfitting} and they do not provide formal guarantees of robustness.

One line of thought pursued by researchers in the field of robustness is finding a method that guarantees robustness and, in other words, provides a mathematical certification. A lot of work has been done in this field \cite{wong2018provable, zhang2018efficient, mirman2018differentiable, weng2018towards, sinha2017certifying, zhang2019towards}, but many of these methods are not applicable to deep neural networks. One of the certified robustness methods applicable to deep neural networks is randomized smoothing. This approach provides a robustness certification by transforming a classifier into a smoothed classifier. By averaging a classifier's prediction on Gaussian-perturbed inputs, randomized smoothing \cite{cohen2019certified} transforms a base classifier into a certifiably robust smoothed classifier.

\begin{figure}[!t]
    \centering 
    \includegraphics[width=1\textwidth]{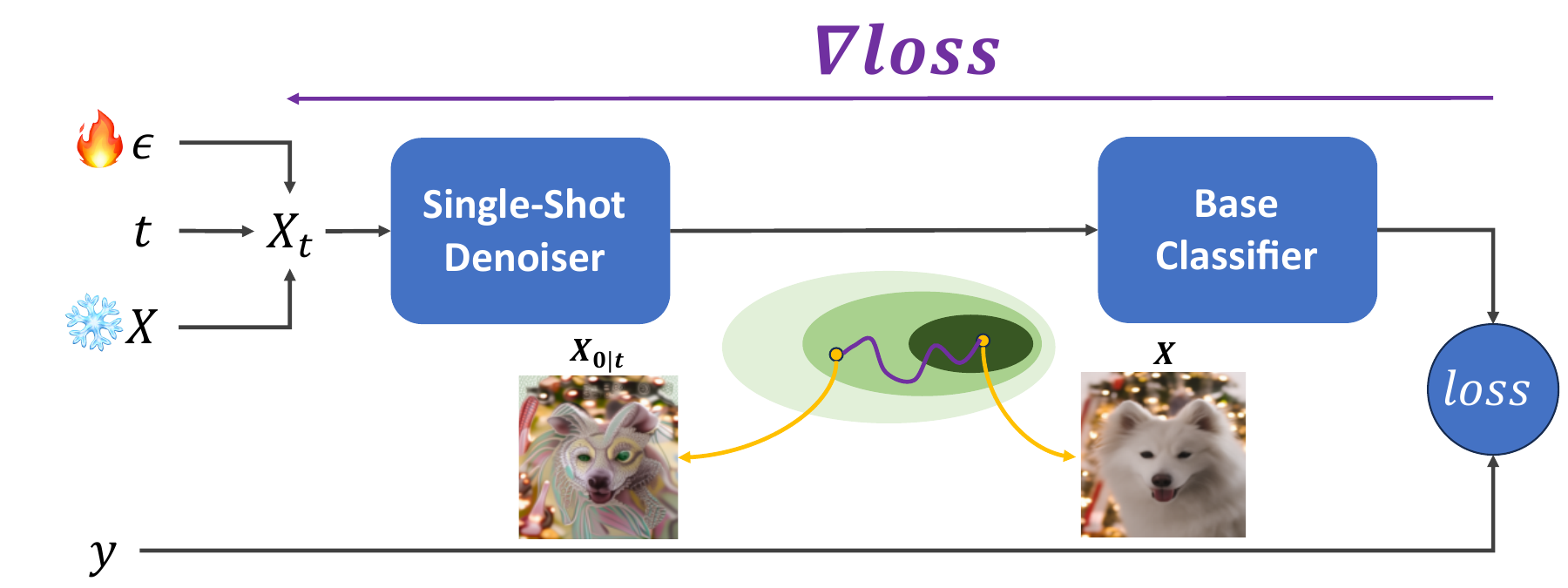}
	
    \caption{Pipeline of our proposed method. We optimize the perturbation $\epsilon$ to induce extreme covariate shift during denoising and utilize it to train the base classifier.
    }
    \label{piple_figure} 
\end{figure}

While randomized smoothing theoretically guarantees robustness of any classifier by transforming it into a smoothed classifier, in practice, we cannot use pretrained classifiers since they are not robust to Gaussian-perturbed data. To address this problem, \cite{salman2020denoised} introduced denoised smoothing, which leverages a denoiser to preprocess perturbed inputs before classification. By incorporating a denoiser, pretrained classifiers can be effectively used within the randomized smoothing paradigm.

One of the denoisers that has achieved excellent results is the denoising diffusion model \cite{ho2020denoising, song2020denoising, song2020score}, which aims to obtain denoised data through a backward process by iteratively denoising the data. \cite{carlini2022certified} demonstrated that by utilizing these models in denoised smoothing, it is possible to achieve excellent results in certifiably robust $l_2$-perturbation using a pre-trained classifier and a pre-trained denoised diffusion model.

However, diffusion-based denoising can introduce a covariate shift when noise estimation errors cause the denoised outputs to deviate significantly from the original clean data distribution. In diffusion-denoised smoothing, this shift propagates into the base classifier’s input, potentially reducing the smoothed classifier’s certified accuracy. Hence, while pretrained classifiers are compatible with diffusion-denoised smoothing, failing to adapt them to this covariate shift results in degraded performance.

To address this issue, we propose a simple yet effective approach to enhance the base classifier’s robustness against covariate shift induced by noise misestimation in diffusion denoised smoothing. Specifically, we introduce a novel adversarial objective function designed around the added noise in denoising diffusion models. Using an adversarial procedure, like projected gradient descent, we identify the noise configuration that generate the largest covariate shift and most effectively confuse the classifier. These adversarial noise samples are then used to train the base classifier, improving its robustness to covariate shift and significantly enhancing the certified performance of the smoothed classifier.

\textbf{Contributions:}
\begin{itemize}
    \item We identify a critical limitation in Diffusion Denoised Smoothing: denoising diffusion models trained with mean squared error (MSE) loss introduce a significant covariate shift due to misestimation of added noise.
    \item To address this issue, we propose a novel adversarial training framework that explicitly targets extreme covariate shift cases, improving classifier robustness under noise misestimation.
    \item We evaluate our approach on three standard classification benchmarks—MNIST, CIFAR-10, and ImageNet—achieving new state-of-the-art results in certified accuracy and Average Certified Radius (ACR).
\end{itemize}

\section{Background}
\label{sec:preliminaries}

\subsection{Denoising Diffusion Models}
The Denoising Diffusion Probabilistic Model (DDPM)\cite{ho2020denoising} is one of the most well-known types of generative models, recognized for its ability to generate high-quality data through a diffusion-based process. DDPM constructs a Markovian chain $\left\{x_t\right\}_{t=0}^T$ with transition probabilities defined as:
\begin{equation}
	p\left(x_t \lvert x_{t-1}\right) = 
	\mathcal{N}\left(
	x_t;
	\sqrt{1 - \beta_t} x_{t-1},
	\sqrt{\beta_t} I
	\right)
\end{equation}
where $\beta_t \in \left(0,1\right)$ controls the amount of noise variance added at each step. Since DDPM follows a simple Gaussian Markov chain, if we define $\bar{\alpha}_t = \prod_{t=1}^t \left(1-\beta_t\right)$, we have:
\begin{equation}
	p\left(x_t \lvert x_{0}\right) = 
	\mathcal{N}\left(
	x_t;
	\sqrt{\bar{\alpha}_t} x_{0},
	\sqrt{1 - \bar{\alpha}_t} I
	\right)
\end{equation}

In other words, $x_t$ can be expressed as:
\begin{equation}
	x_t = \sqrt{\bar{\alpha}_t} x_0 + \sqrt{1 - \bar{\alpha}_t} \epsilon
\end{equation}
where $\epsilon \sim \mathcal{N}\left(0,I\right)$ and it is independent of $x_0$.

In the reverse process, $x_t$ is progressively denoised step by step using a learned variational Markov chain. This chain is modeled as:
\begin{equation}
	p\left(x_{t-1} \lvert x_t\right) = 
	\mathcal{N}\left(
	x_{t-1};
	\mu_{\theta}\left(x_t, t\right),
	\Sigma_{\theta}\left(x_t, t\right)
	\right)
\end{equation}
where $\mu_{\theta}\left(x_t,t\right)$ and $\Sigma_{\theta}\left(x_t,t\right)$ are parameterized by neural networks and learned during training. \cite{ho2020denoising} propose setting $\mu_{\theta}\left(x_t,t\right)$ as:
\begin{equation}
	\mu_{\theta}\left(x_t, t\right) =
	\frac{1}{\sqrt{1 - \beta_t}}\left(
	x_t - \frac{\beta_t}{\sqrt{1 - \bar{\alpha}_t}} \epsilon_{\theta}\left(x_t, t\right)
	\right)
\end{equation}
where $\epsilon_\theta\left(x_t,t\right)$ is a denoiser that predicts the noise added to the data at time step $t$. In DDPM, the covariance matrix $\Sigma_{\theta}\left(x_t,t\right)$ is simplified to $\sigma_t^2 I$, where $\sigma_t^2$ is a fixed variance at each time step $t$, and $I$ is the identity matrix. This simplifies the learning process by assuming an isotropic Gaussian distribution for the noise.

To train this model, the following optimization problem must be solved:
\begin{equation}
	\underset{\theta}{\min}
	\underset{
		\substack{
			x_0 \sim \mathcal{D}^{trian} \\
			t \in U\left(0, T\right) \\
			\epsilon \sim \mathcal{N}\left(0, I\right)
		}
	}{\mathbb{E}} \left[
	\left\|
	\epsilon - \epsilon_{\theta}\left(x_t, t\right)
	\right\|_2^2
	\right]
\end{equation}
where $\epsilon \sim \mathcal{N}\left(0,I\right)$ represents the true noise added to the data, and $\epsilon_{\theta}\left(x_t,t\right)$ is the model's predicted noise. This objective minimizes the difference between the true noise and the predicted noise across time steps.

In the reverse process of DDPM, the following equation is used to denoise and reconstruct the data:
\begin{equation}
	x_{t-1} = \mu_{\theta}\left(x_t, t\right) + \sigma_t z
\end{equation}
where $z \sim \mathcal{N}\left(0,I\right)$ is a random Gaussian noise sample.

\subsection{Randomized Smoothing}
One widely recognized method for achieving certified robustness in deep neural networks is randomized smoothing \cite{cohen2019certified}. In randomized smoothing, a classifier is transformed into a "smoothed classifier," offering a robustness guarantee for the newly constructed model. Let $f:\mathbb{R}^d \rightarrow \mathcal{Y}$ represent a hard classifier that maps an input to its predicted class. The smoothed version of classifier $f$ is then defined as follows:
\begin{equation}
	\hat{f}(x) = \underset{c \in \mathcal{Y}}{\arg \max} \ \mathbb{P}\left[ f\left(x + \delta\right) = c\right]
\end{equation}
\begin{equation*}
	\begin{array}{cc}
		where &
		\delta \sim \mathcal{N}\left(0, \sigma^2I\right)
	\end{array}
\end{equation*}
where $\sigma$ controls robustness/accuracy tradeoff. The following theorem provides an $l_2$ robustness certification for randomized smoothing.

\begin{theorem}[\cite{cohen2019certified}]
    Let $f: \mathbb{R}^{d} \rightarrow \mathcal{Y}$ be any base classifier, let $\sigma > 0$, and let $\hat{f}$ be its associated smooth classifier and we have:
    \begin{equation}
        a = \underset{c \in \mathcal{Y}}{\arg \max} \ \mathbb{P}\left[ f\left(x + \delta\right) = c\right]
    \end{equation}
    and,
    \begin{equation}
        b = \underset{c \in \mathcal{Y} - \left\{a\right\}}{\arg \max} \ \mathbb{P}\left[ f\left(x + \delta\right) = c\right]
    \end{equation}
    to be the two most likely classes for $x$ according to $\hat{f}$ with probability $p_{a}$ and $p_{b}$. Then, we have that $\hat{f}\left(x'\right) = a$ for all $x^{'}$ satisfying
    \begin{equation}
        \|x' - x\|_{2} 
        \leq 
        \frac{\sigma}{2}\left(
            \Phi^{-1}\left(p_a\right) - \Phi^{-1}\left(p_b\right)
        \right)
    \end{equation}
    where $\Phi^{-1}\left(\cdot\right)$ is inverse CDF of standard normal.
\end{theorem}

Although randomized smoothing offers a robustness certification for the smoothed classifier, in practice, the performance of the smoothed classifier tends to be poor because the base classifier is not inherently robust to Gaussian perturbations. \cite{cohen2019certified} address this challenge by training the base classifier with Gaussian data augmentation, improving its robustness to Gaussian perturbations. However, in SmoothAdv \cite{salman2019provably}, it is demonstrated that adversarial training of the base classifier using adversarial examples generated from the smoothed classifier within an $l_2$-ball leads to better performance. SmoothMix \cite{jeong2021smoothmix} introduces a method where an adversarial example is generated from a clean input, and an auxiliary loss is computed using mix-up augmentation between the clean and adversarial examples. It then trains the base classifier using both the auxiliary loss and the Gaussian augmentation loss.

MACER \cite{zhai2020macer} takes a different approach by directly training the smoothed classifier rather than the base classifier. Its training process focuses on maximizing the certified robustness radius of the smoothed classifier. Alternatively, Consistency \cite{jeong2020consistency} uses a distinct method by training the classifier to ensure that the smoothed classifier produces stable outputs within a local neighborhood, while also ensuring that the output remains as far from a uniform distribution as possible.

\subsection{Denoised Smoothing}
Although randomized smoothing offers robustness certification for the smoothed version of any classifier, in practice,it requires training the base classifier on noisy/adversarial examples, making it impractical to use a pre-trained classifier. Denoised Smoothing \cite{salman2020denoised} addresses this challenge by incorporating a denoiser before the base classifier, allowing for the use of pre-trained classifiers while still benefiting from randomized smoothing. Let $f:\mathbb{R}^d\rightarrow\mathcal{Y}$ be a classifier and $D:\mathbb{R}^d\rightarrow\mathbb{R}^d$ be a denoiser. The denoised smoothed version of the classifier $f$ is defined as follows:
\begin{equation}
	g(x) = \underset{c \in \mathcal{Y}}{\arg \max} \ \mathbb{P}\left[ f\left(D\left(x + \delta\right)\right) = c\right]
\end{equation}
\begin{equation*}
	\begin{array}{cc}
		where &
		\delta \sim \mathcal{N}\left(0, \sigma^2I\right)
	\end{array}
\end{equation*}

It can be easily demonstrated that the robustness certification remains valid for denoised smoothed classifiers as well. The denoiser in denoised smoothing can be trained using Mean Squared Error (MSE) loss, stability loss \cite{salman2020denoised}, or a combination of both. \cite{salman2020denoised} demonstrate that training with stability loss leads to better results.

Although a pretrained classifier can be utilized in the denoised smoothing approach, training the denoiser remains essential.
To address this issue, \cite{carlini2022certified} propose Diffusion Denoised Smoothing. In this approach, the noisy input is denoised using a denoising diffusion model at a specific time step $t^\ast$, determined as: 
\begin{equation}
	\label{t_denoising_equation}
	t^\ast= \underset{t}{\arg\min}\ {\left| \sigma - \sqrt{\frac {1-\bar{\alpha}_t}{\bar{\alpha}_t}} \right|}
\end{equation}

This formulation selects the optimal time step $t^\ast$ where the diffusion noise level best matches the target noise parameter $\sigma$ and $x_{t^\ast}=\sqrt{\bar{\alpha}_{t^\ast}} (x+\delta)$. Since the denoising process in denoising diffusion models is iterative, it has a computational complexity of $O(t^\ast)$. This not only slows down the denoising process but also causes error accumulation over multiple steps, leading to a substantial discrepancy between the denoised image and its corresponding clean image. To address this challenge,  \cite{carlini2022certified} propose the use of a single-shot denoiser:
\begin{equation}
		x_{0 \lvert t^\ast} = {\mathbb{E}} \left[
		x_0 | x_{t^\ast} \right] =
		\frac{
		x_{t^\ast} - \sqrt{1 - \bar{\alpha}_{t^\ast}} {\epsilon_\theta}\left(x_{t^\ast}, t^\ast\right)
	}{
		\sqrt{\bar{\alpha}_{t^\ast}}
	}
	\label{eq.11}
\end{equation}

\begin{figure*}[!t]
	\centering 
	\subfloat[$l_2$]{ 
		\label{fig_cifar10_mean_l2_distances}
		\includegraphics[width=0.4\textwidth]{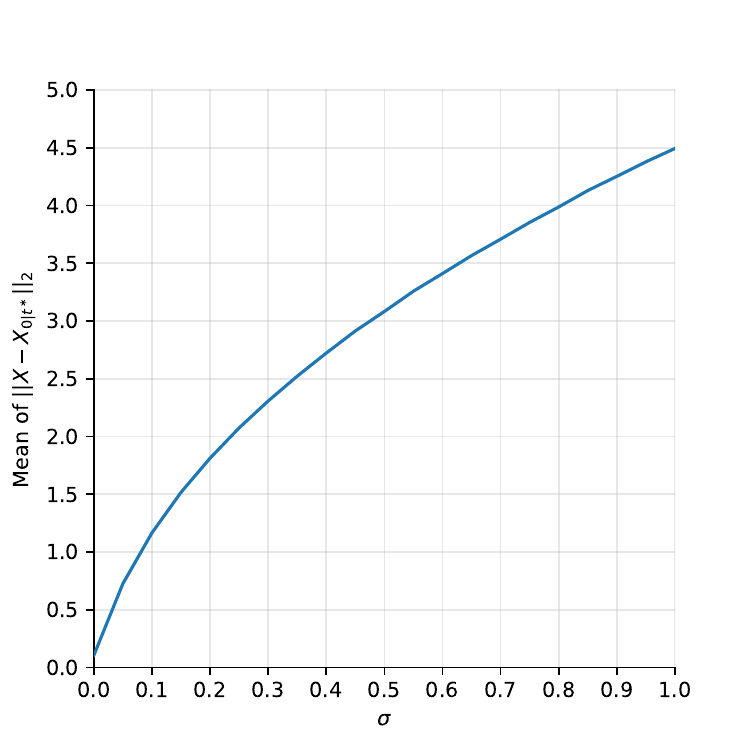}
	}
	\subfloat[LPIPS]{
		\label{fig_cifar10_mean_lpips_distances}
		\includegraphics[width=0.364\textwidth, trim=2mm 2mm 0 0, clip]{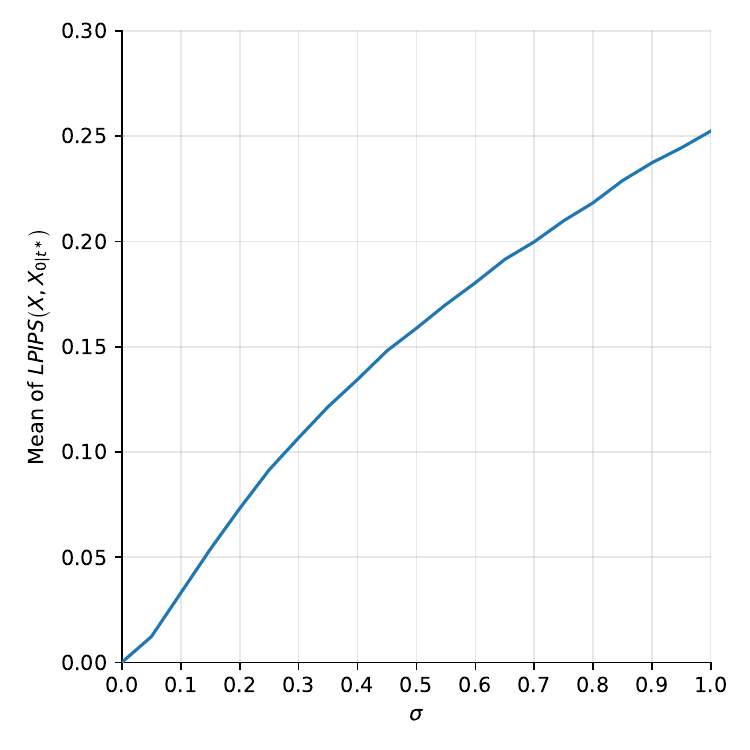}
	}
	\caption{
		Mean (a) $l_2$ and (b) LPIPS distances between clean samples ($x_0$) and noisy-then-denoised samples ($x_{0|t^\ast}$) at varying noise standard deviations. A total of 3,000 samples were randomly selected from the CIFAR-10 dataset. For each sample, the noisy-then-denoised counterpart was computed using the time formulation described in Equation \eqref{t_denoising_equation}.
	}
	\label{fig_l2_and_lpips_distances_vs_sigma} 
\end{figure*}

\section{Method}
\cite{carlini2022certified} assert that Diffusion Denoised Smoothing is versatile, allowing the use of any pre-trained classifier. However, our observations in section \ref{cov_shift_issue_section}, reveal a challenge with single-shot denoisers that may degrade the smoothed classifier performance. To address this challenge, we introduce a novel adversarial objective function to train the base classifier, as discussed in section \ref{cov_shift_solution_section}.

\subsection{Covariate Shift Issue in Single-Shot Denoiser}
\label{cov_shift_issue_section}
The single-shot denoiser in eq.\ref{eq.11} is a key component of Diffusion Denoised Smoothing, allowing it to integrate with any pre-trained base classifier. The effectiveness of this component plays an important role in the smoothed classifier performance. To evaluate the efficacy of the single-shot denoiser, we conducted an experiment on a randomly selected subset of 3000 CIFAR-10 samples. In our experiment, every sample is corrupted with Gaussian noise of a specified standard deviation and subsequently denoised using the single-shot denoiser at time $t^\ast$, as derived from equation \eqref{t_denoising_equation}. The noisy-then-denoised sample is denoted by $X_{0 \lvert t^\ast}$. The efficiency of the single-shot denoiser is measured by calculating the distance between the clean and the noisy-then-denoised sample. The $l_2$ and LPIPS\footnote{Learned Perceptual Image Patch Similarity} \cite{zhang2018unreasonable} distances are chosen to evaluate geometric and perceptual similarity, respectively. Figure \ref{fig_l2_and_lpips_distances_vs_sigma} shows $l_2$ and LPIPS distances across varying $\sigma$ levels, illustrating a clear increase in distances between clean and noisy-then-denoised data as $\sigma$ grows. Significant geometric distances indicate that the noisy data is mapped to a point far from the clean data, potentially outside the base classifier’s decision boundary around the clean data. This can result in changes to the base classifier’s predictions, thereby degrading its performance and, ultimately, the performance of the smoothed classifier. Significant perceptual distances further support this observation.

How can this covariate shift mathematically be justified? For this purpose, we need to examine the denoising diffusion equations. In the forward path, the clean sample is corrupted by a Gaussian noise as follows:
\begin{equation}
	\label{cov_shift_section_x_t_equtation}
	x_t = \sqrt{\bar{\alpha}_t} x + \sqrt{1 - \bar{\alpha}_t} \epsilon
\end{equation}

where $\epsilon$ is a standard Gaussian noise. By denoising this corrupted sample using a single-shot denoiser, we obtain
\begin{equation}
	x_{0 \lvert t} = \frac{
		x_t - \sqrt{1 - \bar{\alpha}_t} \tilde{\epsilon}_{\theta}\left(x_t, t\right)
	}{
		\sqrt{\bar{\alpha}_t}
	}.
\end{equation}
Here, $\tilde{\epsilon}\left(x_t, t\right)$ denotes the denoiser’s estimate of the noise $\epsilon$. Now, By substituting $x_t$ from equation \eqref{cov_shift_section_x_t_equtation}, we have:

\begin{equation}
	x_{0 \lvert t} = x + \frac{
		\sqrt{1 - \bar{\alpha}_t}
	}{\sqrt{\bar{\alpha}_t}} \left(
	\epsilon - \tilde{\epsilon}_{\theta}\left(x_t, t\right)
	\right)
\end{equation}

Let the mismatch term $\epsilon_{t}\left(x_t\right) = \epsilon - \tilde{\epsilon}_{\theta}\left(x_t, t\right)$ quantify the discrepancy between actual noise and its predicted counterpart. then, 
\begin{equation}
	x_{0 \lvert t} = x + \frac{
		\sqrt{1 - \bar{\alpha}_t}
	}{\sqrt{\bar{\alpha}_t}} \epsilon_{t}\left(x_t\right)
\end{equation}
Thus, the noisy-then-denoised sample ($x_{0\lvert t}$) represents the clean sample ($x$), augmented by mismatch term. Since $\epsilon_{t}\left(x_t\right)$ may be non-negligible, the resulting sample $x_{0 \lvert t}$ can lie far from the original distribution, causing the base classifier to operate on out-of-distribution inputs. As this distribution shift occurs in the input space of the base classifier, we refer to it as a \emph{covariate shift} issue.

\subsection{Dealing with Covariate Shift}
\label{cov_shift_solution_section}
\cite{carlini2022certified} draw inspiration from denoised smoothing to develop diffusion denoised smoothing. By incorporating a denoiser before the classifier, denoised smoothing allows randomized smoothing to be effectively applied to any pretrained classifier. In denoised smoothing, the denoiser can be trained using either MSE loss, stability loss, or a combination of both. As discussed in \cite{salman2020denoised}, the stability loss aligns the base classifier's predictions on noisy-then-denoised data with its predictions on clean data. Superior results can be achieved by employing this loss exclusively during the training of the denoiser. Stability loss \cite{salman2020denoised} strengthens the base classifier robustness against covariate shifts that may arise within the denoiser.

Diffusion Denoised Smoothing utilizes a single-shot denoiser derived from a denoising diffusion model. This model has been trained using MSE loss to effectively predict the added noise. Indeed, it appears that utilizing stability loss is more effective in enhancing the performance of diffusion denoised smoothing classifiers.

Thus, the observations in section \ref{cov_shift_issue_section} confirm the presence of covariate shift, and the preceding discussion highlights that employing a loss function like stability loss can be more effective. There are two approaches to address the covariate shift challenge. One approach to eliminate the covariate shift is to ensure that the input and output of the single-shot denoiser are closely aligned. To accomplish this, a diffusion model can be trained using a loss function such as mean squared error (MSE), enabling its corresponding single-shot denoiser accurately predicts the clean sample. By doing so, a denoising diffusion model can be transformed into a consistency model \cite{song2023consistency}, effectively addressing covariate shift. But it is shown stability works better for training the denoiser. However, while stability loss has been shown to perform better for training the denoiser, training a denoising diffusion model with a new loss function remains computationally expensive and not always practical.

We propose an alternative approach that enhances the robustness of the base classifier to the covariate shift introduced by the single-shot denoiser, offering a significantly faster solution compared to training the single-shot denoiser. Since the covariate shift arises from the misestimation of added noise to the clean data, it is essential to ensure that the base classifier is robust to such misestimation. Therefore, the following optimization problem is proposed:
\begin{equation}
	\label{our_cov_shift_solution}
	\underset{\phi}{\min} \underset{
		\substack{
			\left(x, y\right) \sim \mathcal{D} 
		}
	}{\mathbb{E}} \left[
	\underset{\epsilon}{\sup} \ 
	- \log \left(
	f_{\phi}\left(
	x_{0\lvert t^\ast}\left( \epsilon \right)
	\right)
	\right)_{y}
	\right]
\end{equation}

where $\phi$ represents the classifier parameters, and $\epsilon$ denotes the Gaussian noise added during the forward pass of the denoising diffusion model. A significant challenge in this optimization problem lies in computing the supremum with respect to $\epsilon$, which is inherently difficult. Rather than computing the supremum globally, we can instead identify the value of $\epsilon$ that locally maximizes the loss. Thus, after sampling $\epsilon_0 \sim \mathcal{N}\left(0, I\right)$ for each sample $\left(x,y\right)$, we iteratively update $\epsilon_0$ using a procedure similar to projected gradient descent to maximize the classification loss:
\begin{equation}
	\epsilon_{m+1}=\Pi\left(\epsilon_{m}+\eta\cdot \text{sgn}(\nabla_\epsilon-\log \left(f_\phi \left(x_{0\lvert t^\ast}\left(\epsilon\right)\right)\right)_y\right)
\end{equation}
Here, $\eta$ is the step size, $\text{sgn}(\cdot)$ denotes the sign function, and $\Pi$ represents the projection operation. The projection is implemented using a clamp operation to ensure that $\epsilon$ remains within a ball of radius $r_{adv}$ centered at $\epsilon_0$.

$\epsilon_0$ is updated over $M$ steps, resulting in $\epsilon_M$ at the end of the process. This $\epsilon_M$ is then utilized to fine-tune the base classifier.
This approach ensures that the base classifier becomes robust to $\epsilon$ values that induce strong covariate shifts capable of misleading the classifier.

\subsection{Extreme Covariate Shifted Data Finding Algorithm}
\label{appendix_extreme_covariate_shifted_data_finding}
In diffusion-denoised smoothing, perturbed data are denoised using a single-shot denoiser derived from a pretrained denoising diffusion model. However, the output of the single-shot denoiser may deviate significantly from the clean data, particularly under larger perturbations. We refer to this issue as covariate shift, arising from the misestimation of the added noise.

To enhance the performance of diffusion-denoised smoothing, we propose identifying instances with extreme covariate shift and training the base classifier on them. Algorithm \ref{extreme_covariate_shifted_data_finding_algorithm} outlines the procedure for identifying these extreme cases. In this algorithm, the initial noise $\epsilon_{0} \sim \mathcal{N}(0, I)$ is randomly sampled and subsequently updated over M steps. At each step, the data are perturbed using the noise from the previous step, $\epsilon_{m-1}$, and then denoised with the single-shot denoiser. The denoised data are then passed through the base classifier, and $\epsilon$ is updated by ascending the gradient of the classification loss to maximize the loss. Finally, the data are perturbed using $\epsilon_{M}$, denoised with the single-shot denoiser, and incorporated into the training of the base classifier. Figures \ref{imagenet_adv_1_step} and \ref{imagenet_adv_10_steps} illustrate the data obtained from this algorithm for different step sizes, with $M=1$ and $M=10$, respectively.

\subsection{Distinction from Existing Approaches}
\label{method_section_distinction_subsection}
\cite{carlini2022certified} describe the covariate shift introduced by single-shot denoisers as 'hallucination after denoising' and improve the performance of diffusion-denoised smoothing \cite{carlini2022certified} by training the base classifier on data augmented with such covariate shifts. In other words, they address the following optimization problem for the base classifier:
\begin{equation}
	\label{carlini_cov_shift_solution}
	\underset{\phi}{\min} \underset{
		\substack{
			\left(x, y\right) \sim \mathcal{D} \\
			\epsilon \sim \mathcal{N}\left(0, I\right)
		}
	}{\mathbb{E}} \left[
	- \log \left(
	f_{\phi}\left(
	x_{0\lvert t^\ast}\left(\epsilon\right)
	\right)
	\right)_{y}
	\right]
\end{equation}

\begin{algorithm}[!t]
    \caption{Extreme Covariate Shifted Data Finding}
    \label{extreme_covariate_shifted_data_finding_algorithm}
    \begin{algorithmic}
        \REQUIRE $f_{\phi}\left(\cdot\right)$, $\tilde{\epsilon}_{\theta}\left(\cdot, \cdot\right)$, $(x, y) \in \mathcal{D}^{train}$, $\sigma$, $\eta$, $M$, $r_{adv}$
        \STATE $\textcolor{red}{\epsilon_0 \sim \mathcal{N}\left(0, I\right)}$
        \STATE $t^\ast$ is calculated from equation \eqref{t_denoising_equation}
        \FOR {$m=1$ to $M$}
            \STATE $x_{t^\ast} = \sqrt{\bar{\alpha}_{t^\ast}} x + \sqrt{1 - \bar{\alpha}_{t^\ast}} \textcolor{red}{\epsilon_{m-1}}$
            \STATE $x_{0\lvert t^\ast} = \frac{x_{t^\ast} - \sqrt{1 - \bar{\alpha}_{t^\ast}} \tilde{\epsilon}_{\theta}\left(x_{t^\ast}, t^\ast\right)}{\sqrt{\bar{\alpha}_{t^\ast}}}$
            \STATE $loss = CrossEntropy\left(f_{\phi}\left(\textcolor{red}{x_{0\lvert t^\ast}}\right), y\right)$
            \STATE $\epsilon_{m} = \epsilon_{m-1} + \eta \cdot \text{sgn}\left(\nabla_{\textcolor{red}{\epsilon_{m-1}}}loss\right)$
            \STATE $\Delta\epsilon = \epsilon_{m} - \epsilon_0$
            \STATE $\epsilon_m = \epsilon_{0} + \text{Clamp}\left(\Delta\epsilon, r_{adv}\right)$
        \ENDFOR
        \STATE $x_{t} = \sqrt{\bar{\alpha}_{t^\ast}} x + \sqrt{1 - \bar{\alpha}_{t^\ast}} \textcolor{red}{\epsilon_{M}}$
        \STATE $x_{0\lvert t^\ast} = \frac{x_{t^\ast} - \sqrt{1 - \bar{\alpha}_{t^\ast}} \tilde{\epsilon}_{\theta}\left(x_{t^\ast}, t^\ast\right)}{\sqrt{\bar{\alpha}_{t^\ast}}}$
        \STATE \textbf{return} $x_{0\lvert t^\ast}$
    \end{algorithmic}
\end{algorithm}

Although exposing the base classifier to covariate shifts during training can enhance its robustness to some extent, it tends to perform poorly on more challenging cases. How can we justify that training the base classifier with covariate shift-augmented data improves the performance of diffusion-denoised smoothing?
This approach builds upon the stability loss concept, with the key distinction that it uses the true class label instead of the classifier's output as the predicted class.
On the other hand, Applying Jensen’s inequality to the convex $-\log\left(\cdot\right)$ function yields:

\begin{equation}
	\label{jenson_inequality_carlini}
	\underset{
		\left(x, y\right) \sim \mathcal{D}
	}{\mathbb{E}}\left[
	-\log \left(
	\underset{
		\substack{
			\epsilon \sim \mathcal{N}\left(0, I\right)
		}
	}{\mathbb{E}} \left[
	f_{\theta}\left(
	x_{0 \lvert t^{\ast}} \left(\epsilon\right)
	\right)
	\right]
	\right)_{y}
	\right]
\end{equation}
\begin{equation*}
	\leq
	\underset{
		\substack{
			\left(x, y\right) \sim \mathcal{D} \\
			\epsilon \sim \mathcal{N}\left(0, I\right)
		}
	}{\mathbb{E}} \left[
	-\log \left(
	f_{\theta}\left(
	x_{0 \lvert t^{\ast}}\left(\epsilon\right)
	\right)
	\right)_{y}
	\right]
\end{equation*}
The right-hand side of the inequality \eqref{jenson_inequality_carlini} represents the loss function in \eqref{carlini_cov_shift_solution}, while the left-hand side corresponds to the classification loss associated with Diffusion Denoised Smoothing. Thus, the loss function in \eqref{carlini_cov_shift_solution} serves as an upper bound for the diffusion denoised smoothing loss function. By minimizing the loss in \eqref{carlini_cov_shift_solution}, we also minimize the diffusion denoised smoothing loss, ultimately enhancing the smoothed classifier's performance.

However, simply augmenting the training data with randomly sampled noisy-then-denoised inputs fails to address more extreme misestimations of noise. We argue that the classifier should be robust not only to typical covariate shifts but also to the most severe shifts that can arise. Therefore, we propose an adversarial training objective that focuses on these hard cases, ensuring that the classifier generalizes better to all noisy-then-denoised inputs. In our proposed method, we first identify the challenging data, and then the base classifier is trained to become robust on these data. By training the base classifier to perform well on challenging data with significant covariate shifts, we anticipate that it will achieve improved performance when exposed to various noisy-then-denoised inputs. By adopting this approach, our method significantly outperforms the standard covariate shift augmentation training procedure in diffusion-denoised smoothing.

\section{Experiment}

Initially, we compare our approach with the method proposed by \cite{carlini2022certified} and demonstrate that enhancing the robustness of the base classifier to the covariate shift introduced by the denoiser significantly improves the $l_2$-defenses of diffusion denoised smoothing. We conducted experiments on three standard classification datasets: MNIST, CIFAR-10, and ImageNet.


For evaluation and comparison with prior approaches, we employed two common criteria:
\begin{itemize}
	\item \textbf{Approximate certified test accuracy at different radii:} This criterion measures the proportion of data points that are correctly classified by the CERTIFY algorithm without abstaining and have a certified radius greater than r.
	\item \textbf{Average Certified Radius (ACR):} This criterion represents the average certified radius returned by the CERTIFY algorithm for test data points that are correctly classified.
	\begin{equation}
		ACR = \frac{1}{\left|\mathcal{D}_{test}\right|}\sum_{\left(x,y\right)\in\mathcal{D}_{test}}{CR\left(f, \sigma, x\right) \mathbf{1}_{\hat{f}\left(x\right)=y}}
	\end{equation}
	where $\mathcal{D}_{test}$ denotes the test dataset, $CR\left(f,\sigma,x\right)$ represents the certified radius returned by the CERTIFY algorithm, and $\hat{f}\left(x\right)$ is the smoothed classifier corresponding to $f$.
\end{itemize}
Approximate certified accuracy gauges performance at specific robustness thresholds, while ACR indicates the overall robustness across the entire test set. Following \cite{cohen2019certified}, we use the same sampling and confidence parameters in the CERTIFY procedure for all models to ensure consistent comparisons.

\textbf{MNIST Configuration.} For the MNIST dataset, we use a pretrained denoising diffusion model available on Hugging Face\footnote{\url{https://huggingface.co/1aurent/ddpm-mnist}} as the denoiser and a LeNet \cite{lecun1998gradient} model as the base classifier. For this dataset, we evaluate the performance with $\sigma = \left\{0.25, 0.50, 1.00\right\}$.

\textbf{CIFAR-10 Configuration.} For image denoising, we use the denoising diffusion model from \cite{nichol2021improved}, following the approach of \cite{carlini2022certified}. To ensure a fair comparison between our method and previous approaches, we employ a ResNet-110 as the base classifier for all algorithms utilizing randomized smoothing for certified robustness. For this dataset, we evaluate the performance with $\sigma = \left\{0.25, 0.50, 1.00\right\}$.

\textbf{ImageNet Configuration.} We denoise ImageNet data with the 552M-parameters denoising diffusion model \cite{dhariwal2021diffusion}, and employ 87M-parameters BEiT-v2 \cite{peng2022beit} base model using implementation from timm \cite{rw2019timm}. For this dataset, we evaluate the performance with $\sigma =\left\{0.5, 1.0\right\}$.

For all datasets, we set $M=1$ unless stated otherwise. The step size is set to $\eta=0.1$ for MNIST and CIFAR-10, and $\eta=0.05$ for ImageNet.

\subsection{CERTIFY and PREDICT Algorithms}
\label{appendix_certify_algorithm_parameters}
By employing a base classifier that is robust to the covariate shift caused by the denoiser, we utilize the same certification and prediction algorithms as in diffusion denoised smoothing \cite{carlini2022certified}. In both algorithms, the input data, which is made noisy using different Gaussian noise samples, is first denoised and then passed to the base classifier.
In the PREDICT algorithm, a majority vote is taken from the obtained outputs. If the number of votes passes a binomial test, the majority vote is returned; otherwise, the algorithm outputs 'abstain'. In the CERTIFY algorithm, we ensure that the majority vote is robust if its proportion exceeds that of the second-highest vote by a specified margin. The CERTIFY algorithm not only provides the classification output but also returns the corresponding certified radius. 

To evaluate the certified accuracy and radii of our method and diffusion denoised smoothing on MNIST, CIFAR-10, and ImageNet, we utilize the diffusion denoised smoothing implementation. For all datasets, the parameters are set as $\alpha=0.001$, $N_0=100$, and $N=10,000$. The CERTIFY algorithm is applied to 500 randomly selected samples from MNIST and CIFAR-10, and 100 randomly selected samples from ImageNet.

\subsection{Results on MNIST}

In this subsection, we examine the performance of our method on MNIST and compare it against existing smoothing-based approaches, emphasizing certified accuracy at various radii and the impact of covariate shift.

Randomized smoothing demonstrates strong performance at smaller radii; however, its efficacy significantly declines as the radius increases. To address this issue, methods such as SmoothAdv \cite{salman2019provably}, MACER \cite{zhai2020macer}, SmoothMix \cite{jeong2021smoothmix}, and Consistency \cite{jeong2020consistency} have been proposed, which aim to directly or indirectly train the smoothed classifier for improved performance on the classification task. By employing these approaches, these methods improve the certified accuracy of the smoothed classifier and reduce the accuracy degradation, compared to randomized smoothing, as the radius increases.

Another line of research focuses on denoised smoothing, which incorporates a denoiser before the base classifier in the smoothed classifier pipeline. By denoising the perturbed data, these methods enable the use of a pretrained classifier as the base classifier.When an effective denoiser, such as a denoising diffusion model, is employed within this framework, the denoised data closely approximates the clean data. However, any residual distance between the denoised data and the clean data can degrade the performance of the base classifier, thereby affecting the overall effectiveness of the smoothed classifier.

\begin{table}[t]
	\caption{Certified top-1 accuracy on MNIST for randomized smoothing and denoised smoothing-based methods. For randomized smoothing methods, the base classifier (LeNet) is trained according to the procedure outlined in each method. In denoised smoothing, a pretrained base classifier (LeNet) on MNIST is used, while the denoiser is trained. In diffusion denoised smoothing (DDS), we compare a pretrained base classifier with a base classifier fine-tuned on covariate-shift-augmented data. Our method fine-tunes the base classifier using the proposed adversarial objective function and employs the resulting classifier in diffusion-denoised smoothing. Each entry shows the certified accuracy, with the clean accuracy for the corresponding model in parentheses, using values reported in the respective papers.}
	\label{mnist_results_table}
	\centering
        \resizebox{\textwidth}{!}{
    	\begin{tabular}{l|ccccc}
                \multicolumn{1}{l}{}& \multicolumn{5}{c}{Certified Accuracy at $\epsilon\left(\%\right)$}\\ \cline{2-6}
    		\multicolumn{1}{l}{Method (MNIST)}
    		& 0.25 & 0.50 & 0.75 & 1.00 &1.25 \\
    		\midrule
    		Randomized Smoothing \cite{cohen2019certified}&$94.6_{(97.4)}$ &$90.2_{(97.4)}$& $82.6_{(97.4)}$& $70.6_{(97.4)}$ & $50.6_{(97.4)}$
    		\\
    		SmoothAdv \cite{salman2019provably}&$96.2_{(98.6)}$ &$92.4_{(97.6)}$& $87.0_{(97.6)}$& $79.2_{(97.6)}$& $59.6_{(97.6)}$\\
    		
    		MACER \cite{zhai2020macer} &$\mathbf{99.0_{(99.0)}}$ &$97.0_{(99.0)}$& $95.0_{(99.0)}$& $90.0_{(99.0)}$& $83.0_{(99.0)}$\\
    		
    		SmoothMix \cite{jeong2021smoothmix}&$\mathbf{99.0_{(99.4)}}$ &$\mathbf{98.2_{(99.4)}}$& $\mathbf{96.9_{(99.4)}}$& $\mathbf{92.7_{(98.8)}}$ & $88.3_{(98.8)}$\\
    		
    		Consistency \cite{jeong2020consistency}&$\mathbf{98.9_{(99.5)}}$ &$\mathbf{98.0_{(99.5)}}$& $96.0_{(99.5)}$& $\mathbf{93.0_{(99.2)}}$& $87.8_{(99.2)}$\\

    		Denoised Smoothing \cite{salman2020denoised}&$92.4_{(94.2)}$ &$90.0_{(94.2)}$& $85.8_{(93.4)}$& $80.0_{(93.4)}$& $71.2_{(93.4)}$\\
    		
    		DDS (Pretrained) \cite{carlini2022certified}&$93.4_{(93.8)}$ &$92.4_{(94.2)}$& $91.4_{(93.8)}$& $90.4_{(94.2)}$& $89.0_{(93.8)}$\\
    		DDS (Finetuned) \cite{carlini2022certified}&$94.2_{(95.0)}$ &$93.6_{(95.0)}$& $92.8_{(95.0)}$& $91.2_{(95.0)}$& $89.8_{(95.0)}$\\
    		\midrule
    		Ours ($r_{adv}=+\infty$, $M=1$) &$95.4_{(96.0)}$ &$94.6_{(96.0)}$& $93.4_{(96.0)}$& $\mathbf{92.8_{(96.0)}}$& $90.4_{(96.0)}$\\
                Ours ($r_{adv}=+0.1$, $M=1$) & $94.8_{(95.4)}$ & $93.8_{(95.4)}$ & $92.6_{(95.0)}$ & $91.8_{(95.0)}$ & $90.2_{(95.0)}$ \\
                Ours ($r_{adv}=+\infty$, $M=2$) & $94.6_{(95.4)}$ & $93.6_{(95.4)}$ & $93.2_{(95.4)}$ & $\mathbf{93.0_{(95.4)}}$ & $\mathbf{92.0_{(95.4)}}$ \\
                Ours ($r_{adv}=+\infty$, $M=4$) & $94.8_{(95.6)}$ & $94.2_{(95.6)}$ & $93.8_{(95.6)}$ & $\mathbf{93.0_{(95.6)}}$ & $91.0_{(95.6)}$ \\
    		
    		\bottomrule
    	\end{tabular}
        }
\end{table}

\begin{figure*}[t]
	\centering 
	\subfloat[$\sigma=0.25$]{ 
		\label{fig_minst_sigma_25}
		\includegraphics[width=0.33\textwidth]{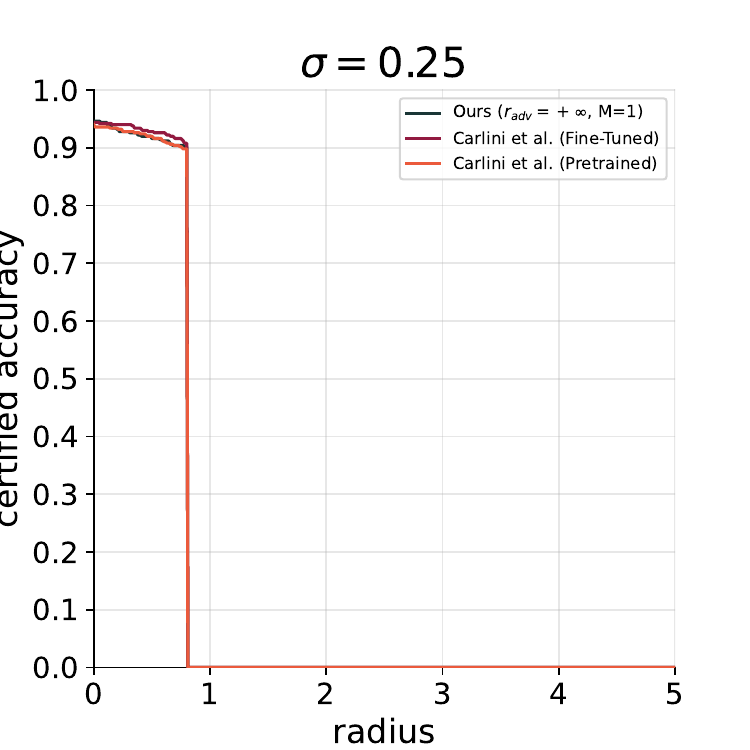}
	}
	\subfloat[$\sigma=0.50$]{ 
		\label{fig_minst_sigma_50}
		\includegraphics[width=0.33\textwidth]{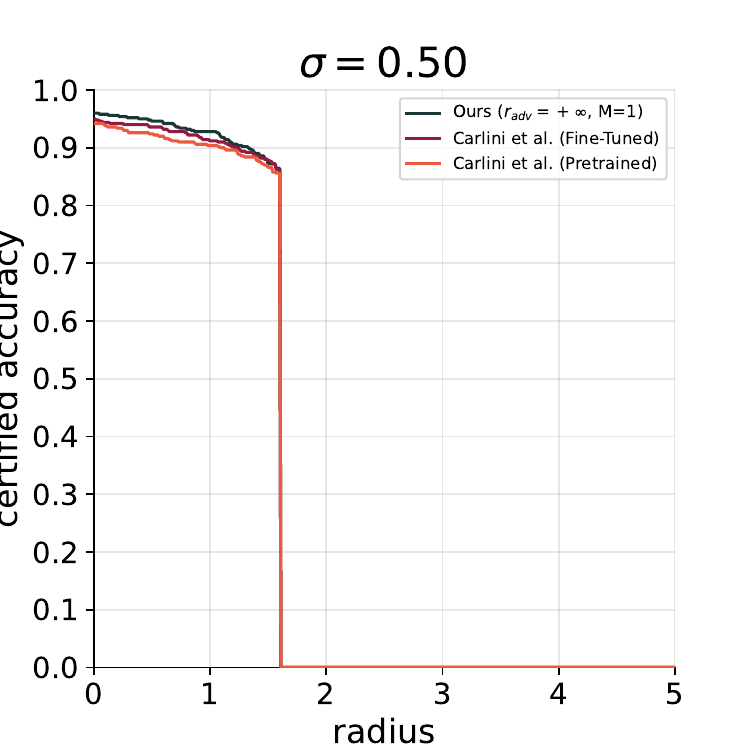}
	}
	\subfloat[$\sigma=1.00$]{ 
		\label{fig_minst_sigma_100}
		\includegraphics[width=0.33\textwidth]{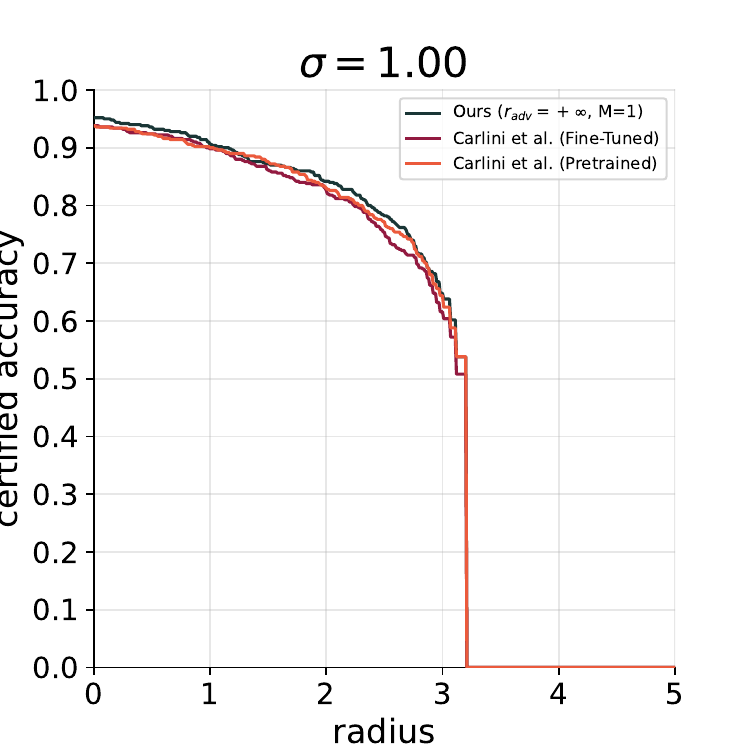}
	}
	\caption{
            Certified accuracy as a function of the $l_2$ adversarial perturbation bound for (a) $\sigma=0.25$, (b) $\sigma=0.50$, and (c) $\sigma=1.00$ on the MNIST dataset.
	}
	\label{fig_mnist_sigmas} 
\end{figure*}

Diffusion denoised smoothing builds upon the denoised smoothing framework by utilizing a denoising diffusion model as the denoiser. As shown in table \ref{mnist_results_table}, This method exhibits poorer performance at smaller radii ($r\leq1.00$) compared to denoiserless approaches, such as SmoothMix \cite{jeong2021smoothmix} and Consistency \cite{jeong2020consistency}, which do not rely on a denoiser. The strong performance of denoiserless methods at smaller radii is understandable, as they train the smoothed classifier directly or indirectly. However, the robustness of denoiserless methods is lower, and their certified accuracy deteriorates much more rapidly compared to diffusion denoised smoothing. As discussed in Appendix~\ref{appendix_single_shot_analysis}, the denoiser maps the perturbed input to a point close to the clean data distribution, on which the base classifier was originally trained. Thus, employing denoised smoothing-based methods becomes more effective at larger radii.

However, 
at larger radii, noise misestimation by the diffusion model can introduce significant covariate shift, leading to a decline in the smoothed classifier's performance. To address this, our proposed method leverages adversarial training on the most challenging misestimated noise scenarios, thereby aligning the smoothed classifier more effectively with the denoiser’s output distribution and improving certified accuracy. As shown in Table \ref{mnist_results_table}, for radii greater than 1.0, utilizing the base classifier trained with our proposed objective significantly improves certified accuracy, achieving state-of-the-art performance and specifically, at $r=1.25$, our approach outperforms the best baseline by $2.2$ percentage points, increasing certified accuracy from $89.8\%$ to $92.0\%$. Table \ref{approximate_certified_test_accuracy_table_mnist} reports the certified accuracy of the smoothed classifier for various values of $\sigma$ used in our experiments. We observe that for $\sigma = 0.5$ and $\sigma = 1.0$, at radii larger than $1.0$, our method achieves state-of-the-art performance on the MNIST dataset. Our method improves the Average Certified Radius (ACR) over diffusion-denoised smoothing, achieving a state-of-the-art ACR of $2.886$ at $\sigma = 1.0$.

Figure \ref{fig_mnist_sigmas} illustrates the certified accuracy of our proposed method as a function of the $l_2$  perturbation radius for different values of $\sigma$ on the MNIST dataset. As $\sigma$ increases, our method demonstrates improved certified accuracy. This improvement can be attributed to greater noise misestimation at higher noise levels, where making the base classifier robust to such misestimation results in better certification performance.


\subsection{Results on CIFAR-10}

\begin{table*}[t]
	\caption{Certified accuracy on CIFAR-10 for randomized smoothing and denoised smoothing-based methods. The columns are defined as in Table 1.}
	\label{results_cifar10_table}
	\centering
        \resizebox{\textwidth}{!}{ 
    	\begin{tabular}{l|cccc}
                \multicolumn{1}{l}{}& \multicolumn{4}{c}{Certified Accuracy at $\epsilon\left(\%\right)$}\\ \cline{2-5}
    		\multicolumn{1}{l}{Method (CIFAR-10)}
    		& 0.25 & 0.50 & 0.75 & 1.00  \\
    		\midrule
    		Randomized Smoothing \cite{cohen2019certified}&$61.0_{(75.0)}$ &$43.0_{(75.0)}$& $32.0_{(65.0)}$& $22.0_{(66.0)}$
    		\\
    		SmoothAdv \cite{salman2019provably}&$\mathbf{74.9_{(84.3)}}$ &$\mathbf{63.4_{(80.1)}}$& $51.9_{(80.1)}$& $39.6_{(62.2)}$\\
    		
    		MACER \cite{zhai2020macer} &$71.0_{(81.0)}$ &$59.0_{(81.0)}$& $46.0_{(66.0)}$& $38.0_{(66.0)}$\\
    		
    		SmoothMix \cite{jeong2021smoothmix}&$67.9_{(77.1)}$ &$57.9_{(77.1)}$& $47.7_{(74.2)}$& $37.2_{(61.8)}$ \\
    		
    		Consistency \cite{jeong2020consistency}&$68.8_{(77.8)}$ &$58.1_{(75.8)}$& $48.5_{(72.9)}$& $37.8_{(52.3)}$\\

    		Denoised Smoothing \cite{salman2020denoised}&$56.0_{(72.0)}$ &$41.0_{(62.0)}$& $28.0_{(62.0)}$& $19.0_{(44.0)}$\\
    		
    		DDS (Pretrained) \cite{carlini2022certified}&$61.4_{(78.4)}$ &$45.2_{(78.4)}$& $30.6_{(62.0)}$& $23.0_{(62.0)}$ \\
    		DDS (Finetuned) \cite{carlini2022certified}&$69.6_{(79.0)}$ &$57.0_{(79.0)}$& $48.8_{(72.8)}$& $42.8_{(72.8)}$\\
    		\midrule
    		Ours&$69.6_{(75.2)}$ &$62.8_{(75.2)}$& $\mathbf{54.6_{(75.2)}}$& $\mathbf{46.6_{(75.2)}}$\\
    		\bottomrule
    	\end{tabular}
        }
\end{table*}

\begin{figure*}[t]
	\centering 
	\subfloat[$\sigma=0.25$]{ 
		\label{fig_cifar10_sigma_25}
		\includegraphics[width=0.33\textwidth]{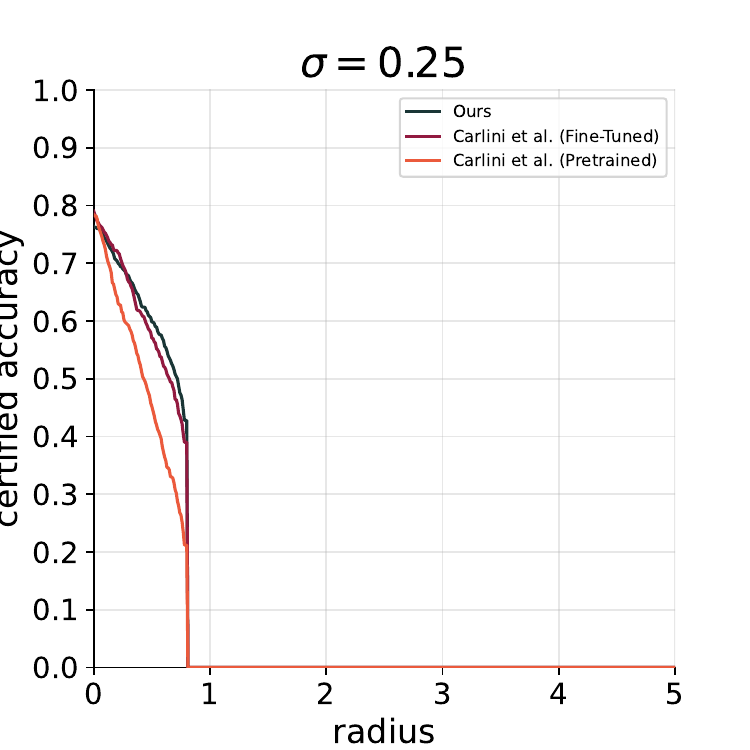}
	}
	\subfloat[$\sigma=0.50$]{ 
		\label{fig_cifar10_sigma_50}
		\includegraphics[width=0.33\textwidth]{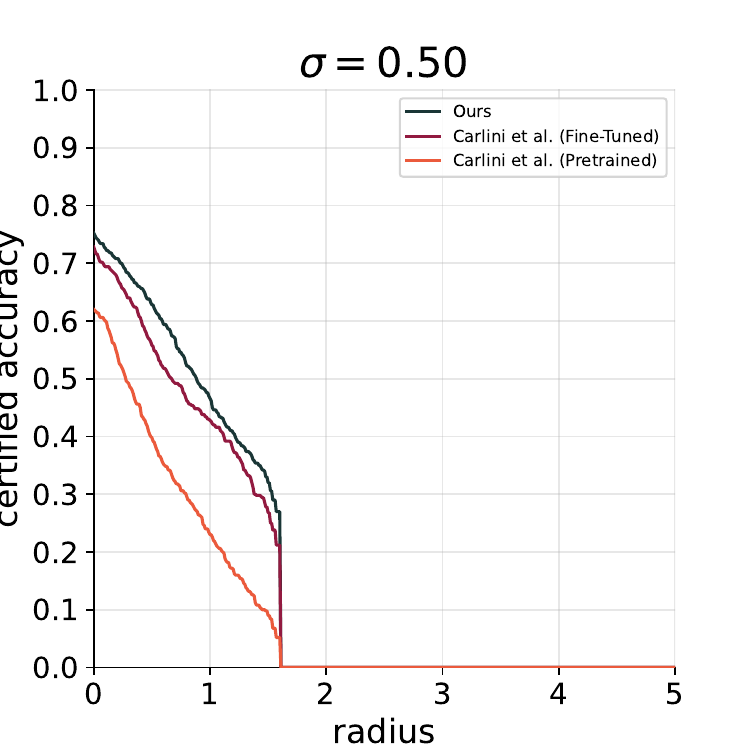}
	}
	\subfloat[$\sigma=1.00$]{ 
		\label{fig_cifar10_sigma_100}
		\includegraphics[width=0.33\textwidth]{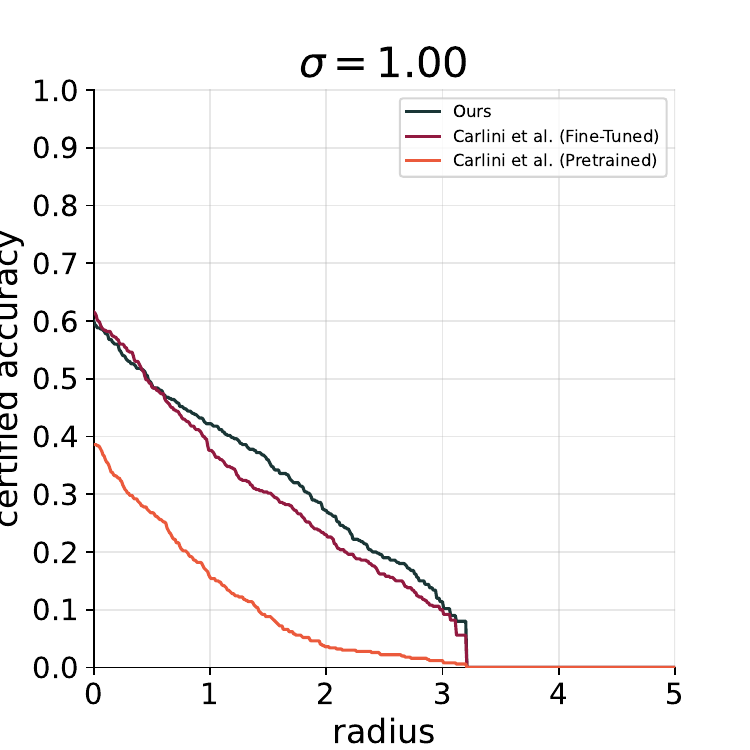}
	}
	\caption{
            Certified accuracy as a function of the $l_2$ adversarial perturbation bound for (a) $\sigma=0.25$, (b) $\sigma=0.50$, and (c) $\sigma=1.00$ on the CIFAR-10 dataset.
	}
	\label{fig_cifar10_sigmas} 
\end{figure*}

Table \ref{results_cifar10_table} illustrates the certified accuracy of various methods on CIFAR-10 across different radii. At smaller radii ($r\leq0.50$), denoiserless methods demonstrate superior performance. For instance, SmoothAdv \cite{salman2019provably} achieves state-of-the-art performance at radii less than $0.50$. This method leverages randomized smoothing while training the base classifier with adversarial data generated from the smoothed classifier. By doing so, this method effectively addresses the classification task for the smoothed classifier to some extent. As a result, it demonstrates better performance at smaller radii. However, as perturbations grow, denoiserless methods degrade more rapidly than diffusion denoised smoothing, indicating their limitations in handling larger perturbations.

Although diffusion denoised smoothing \cite{carlini2022certified} allows the use of any pretrained classifier, its performance at radii less than 1.00 is inferior compared to other methods. its certified accuracy is significantly lower than denoiserless methods at radii below $1.00$. Specifically, at $r=0.75$, it lags $21.3$ percentage points behind the state-of-the-art, and at $r=1.00$, the gap remains $16.6$ percentage points. This is due to the covariate shift caused by the misestimation of the added perturbation during the denoising process. \cite{carlini2022certified} address this issue by training the base classifier on covariate shift-augmented data. By doing so, they achieve state-of-the-art performance at a radius of 1.00; however, their certified accuracy is 3.1 percentage points lower than the state-of-the-art at a radius of 0.75.

As mentioned in section \ref{method_section_distinction_subsection}, while training the base classifier on covariate shift-augmented data can enhance the performance of diffusion denoised smoothing, it fails to make the classifier robust against extreme misestimations of noise. Thus, in our proposed method, we first identify these extreme cases and then train the base classifier on them to enhance its robustness against such scenarios. Table\ref{results_cifar10_table} shows that training the base classifier with our proposed adversarial objective improves certified accuracy by $2.7\%$ and $3.8\%$ at radii 0.75 and 1.00, respectively, achieving state-of-the-art performance at these radii. Table \ref{approximate_certified_test_accuracy_table_cifar10} shows that applying our proposed method increases the Average Certified Radius (ACR) of diffusion denoised smoothing for $\sigma = 0.25$, $\sigma = 0.50$, and $\sigma = 1.00$.

Figure \ref{fig_cifar10_sigmas} compares the certified accuracy of our proposed method with denoised diffusion smoothing as a function of the $l_2$ perturbation radius for various values of $\sigma$ on the CIFAR-10 dataset. As illustrated, improving the robustness of the base classifier against noise misestimation plays a crucial role in enhancing certified accuracy. Since our method explicitly trains the base classifier to be robust against the specific distribution of noise misestimation, it consistently outperforms the baseline approach that uses random noise for robustness.

\subsection{Results on ImageNet}
\begin{table*}[t]
	\caption{Certified accuracy on ImageNet for methods based on diffusion denoised smoothing. The columns are defined as in Table 1.}
	\label{results_imagenet_table}
	\centering
        \resizebox{\textwidth}{!}{
    	\begin{tabular}{l|cccccc}
                \multicolumn{1}{l}{}& \multicolumn{6}{c}{Certified Accuracy at $\epsilon\left(\%\right)$}\\ \cline{2-7}
    		\multicolumn{1}{l}{Method (ImageNet)}
    		& 0.5 & 1.0 & 1.5 & 2.0 & 2.5 & 3.0  \\
    		\midrule		
    		DDS (Pretrained) \cite{carlini2022certified}&$62_{(73)}$ &$54_{(73)}$& $40_{(73)}$& $30_{(59)}$& $23_{(59)}$ &$13_{(59)}$\\
    		DDS (Finetuned) \cite{carlini2022certified}&$68_{(76)}$ &$59_{(76)}$& $49_{(76)}$& $39_{(67)}$& $32_{(67)}$ & $22_{(67)}$\\
    		\midrule
    		Ours&$\mathbf{74_{(79)}}$ &$\mathbf{64_{(79)}}$& $\mathbf{56_{(79)}}$& $\mathbf{40_{(68)}}$& $\mathbf{38_{(68)}}$ & $\mathbf{29_{(68)}}$\\
    		\bottomrule
    	\end{tabular}
        }
\end{table*}

\begin{figure*}[!t]
	\centering 
	\subfloat[$\sigma=0.50$]{ 
		\label{fig_imagenet_sigma_50}
		\includegraphics[width=0.33\textwidth]{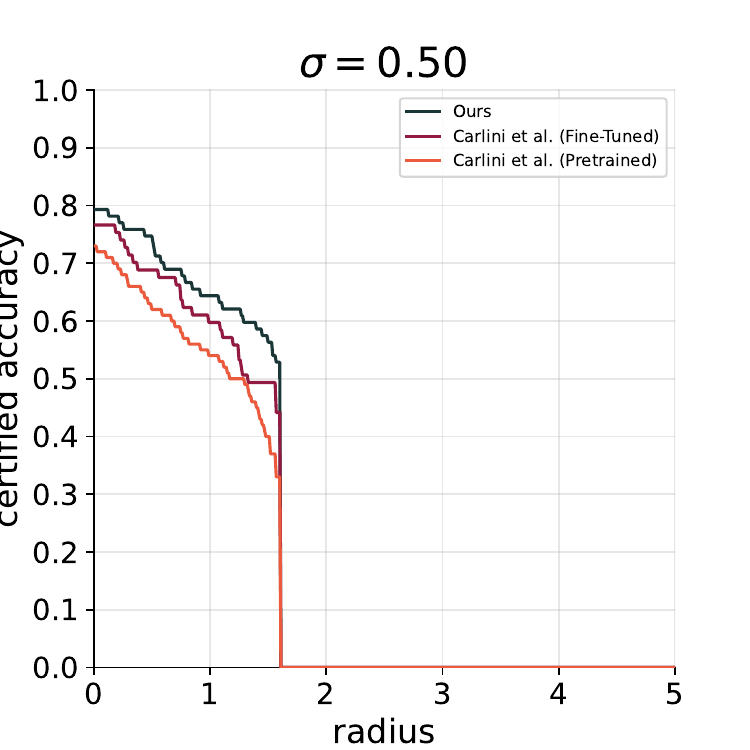}
	}
	\subfloat[$\sigma=1.00$]{ 
		\label{fig_imagenet_sigma_100}
		\includegraphics[width=0.33\textwidth]{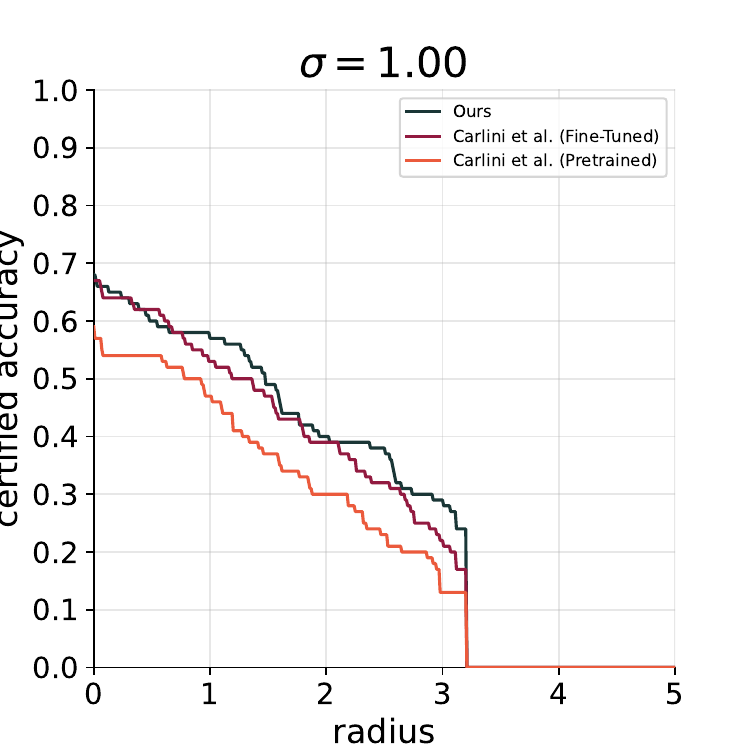}
	}
	\caption{
            Certified accuracy as a function of the $l_2$ adversarial perturbation bound for (a) $\sigma=0.50$, and (b) $\sigma=1.00$ on the ImageNet dataset.
	}
	\label{fig_imagenet_sigmas} 
\end{figure*}
To assess our method in a more challenging setting, we use ImageNet as a benchmark to compare it with diffusion denoised smoothing. Table \ref{results_imagenet_table} compares the performance of our method with those of the two methods proposed by \cite{carlini2022certified}. As observed, training the base classifier with our proposed objective outperforms training with covariate shift-augmented data. This is because the single-shot denoiser is more prone to noise misestimation in high-dimensional data, leading to greater covariate shift. Therefore, it is preferable to enhance the robustness of the base classifier on extremely covariate-shifted data. Figure \ref{imagenet_adv_1_step} presents examples of the extreme covariate-shifted data identified by our method. As observed, the covariate shift can become more pronounced.

By training the base classifier with our proposed adversarial objective, we achieve state-of-the-art performance across all radii. Table \ref{approximate_certified_test_accuracy_table_imagenet} shows that our method improves the state-of-the-art ACR, increasing it from $1.311$ to $1.367$ for $\sigma=0.5$ and from $2.113$ to $2.212$ for $\sigma=1.0$, demonstrating an overall improvement in robustness across the entire test set.

As shown in Figure \ref{fig_imagenet_sigmas}, similar performance trends are observed on the high-dimensional ImageNet dataset. Our method consistently outperforms the baseline denoised diffusion smoothing approach across different noise levels, highlighting the advantage of explicitly training the base classifier to handle noise misestimation even in more complex and large-scale settings.

\subsection{Analyzing Hyperparameters}
Our proposed method introduces three hyperparameters: $\eta$, $M$, and $r_{adv}$. Below, we analyze the impact of each hyperparameter and discuss how to select the optimal setting.

The parameter $\eta$ represents the step size used in the optimization process. In each row of Figures \ref{imagenet_adv_1_step} and \ref{imagenet_adv_10_steps}, increasing the step size amplifies the covariate shift observed in the image. Therefore, this parameter should be chosen such that the induced covariate shift does not significantly alter the image content. Empirically, we find that setting $\eta=0.1$ yields favorable results.

The radius within which $\epsilon$ is updated is controlled by $r_{adv}$. Our observations on the MNIST dataset (Table \ref{approximate_certified_test_accuracy_table_mnist}) suggest that optimization performs best when the radius is unrestricted, i.e., setting $r_{adv} = +\infty$.

Another parameter in our method is $M$, which represents the number of update steps for $\epsilon$. Figures \ref{imagenet_adv_1_step} and \ref{imagenet_adv_10_steps} illustrate that increasing $M$ from $M=1$ to $M=10$—while keeping the step size constant—amplifies the covariate shift. Table \ref{approximate_certified_test_accuracy_table_mnist} shows that increasing $M$ reduces certified accuracy at small radii but improves it at larger radii. This occurs because training the base classifier with a larger $M$ exposes it to more covariate-shifted data, enhancing its robustness to covariate shift. As the radius increases, the covariate shift also intensifies. Consequently, base classifiers trained with larger $M$ exhibit greater robustness to covariate shift, leading to improved performance of the corresponding smoothed classifier at larger radii. Although increasing the number of update steps for $\epsilon$ enhances the performance of the smoothed classifier at larger radii, it significantly increases training time. Therefore, we set $M=1$ for experiments on CIFAR-10 and ImageNet, demonstrating the effectiveness of our proposed method.

\section{Conclusion}
Diffusion Denoised Smoothing extends randomized smoothing by enabling the use of pretrained classifiers, demonstrating strong robustness at larger radii compared to denoiserless approaches. However, its performance is hindered by covariate shift caused by noise misestimation during the denoising. \cite{carlini2022certified} handle this shift by augmenting training with covariate shift-affected data, but their approach fails to address extreme misestimations. To overcome this limitation, we propose a novel adversarial training objective that explicitly identifies and trains on the most challenging covariate shift cases. Our method significantly improves certified accuracy and Average Certified Radius (ACR), achieving new state-of-the-art performance in diffusion denoised smoothing.

\bibliography{sample}

\clearpage
\appendix
\section{Single-Shot Denoiser Analysis}
\label{appendix_single_shot_analysis}
To gain a deeper understanding of single-shot denoisers, we analyze their performance on 3,000 randomly selected samples from the CIFAR-10 dataset. Each sample is perturbed with Gaussian noise of a specified standard deviation and subsequently denoised using the single-shot denoiser with the time parameter obtained from Equation \eqref{t_denoising_equation}.

\begin{figure*}[!b]
    \centering 
    \includegraphics[width=1\textwidth, trim=0mm 50mm 0mm 50mm, clip]{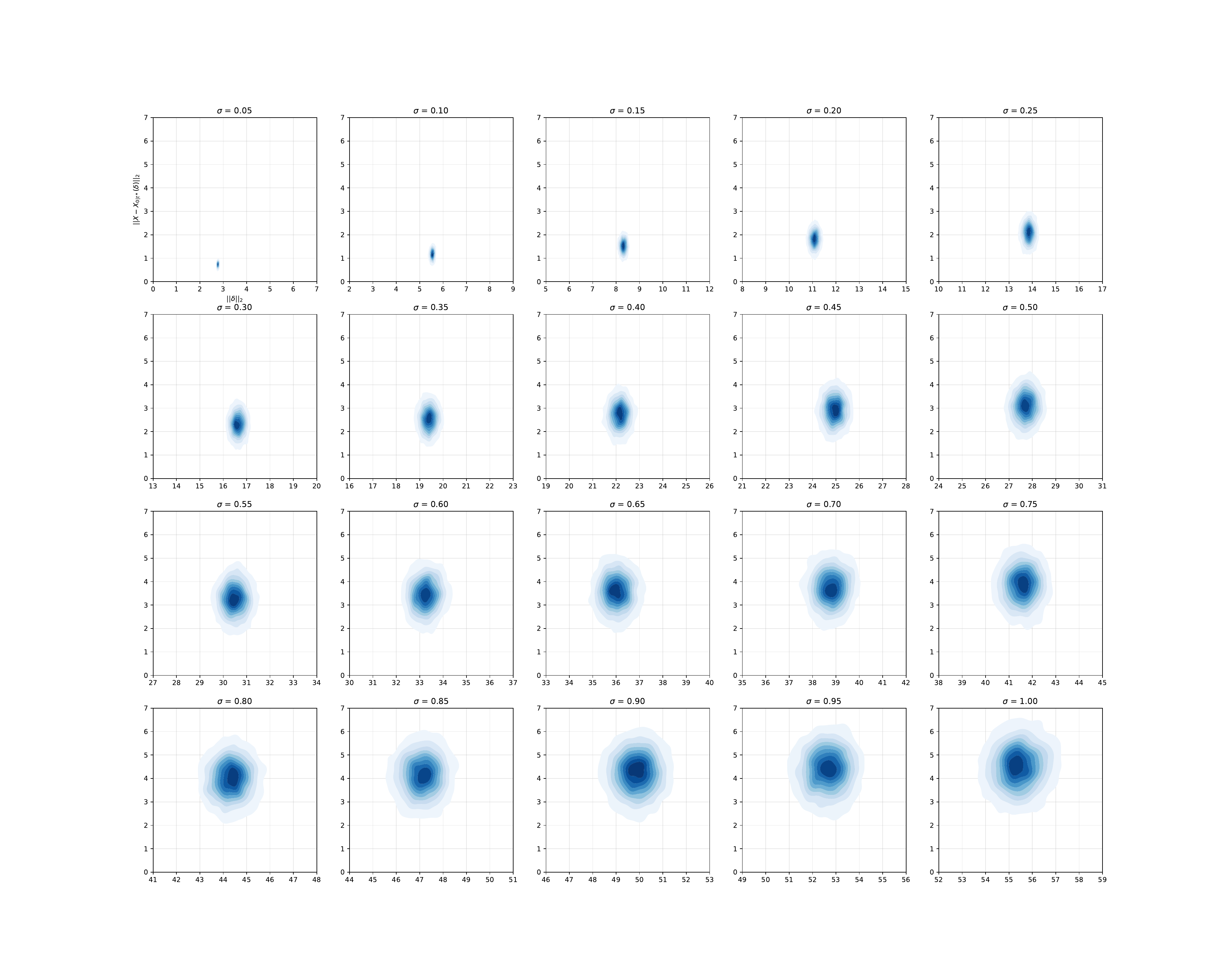}
	
    \caption{
        KDE plot of the $l_2$ distance between clean data and the base classifier's input for randomized smoothing methods without a denoiser (x-axis) and the corresponding distance for diffusion-denoised smoothing (y-axis).
    }
    \label{cifar10_l2_distances} 
\end{figure*}

\begin{figure*}[!b]
    \centering 
    \includegraphics[width=1\textwidth, trim=0mm 50mm 0mm 50mm, clip]{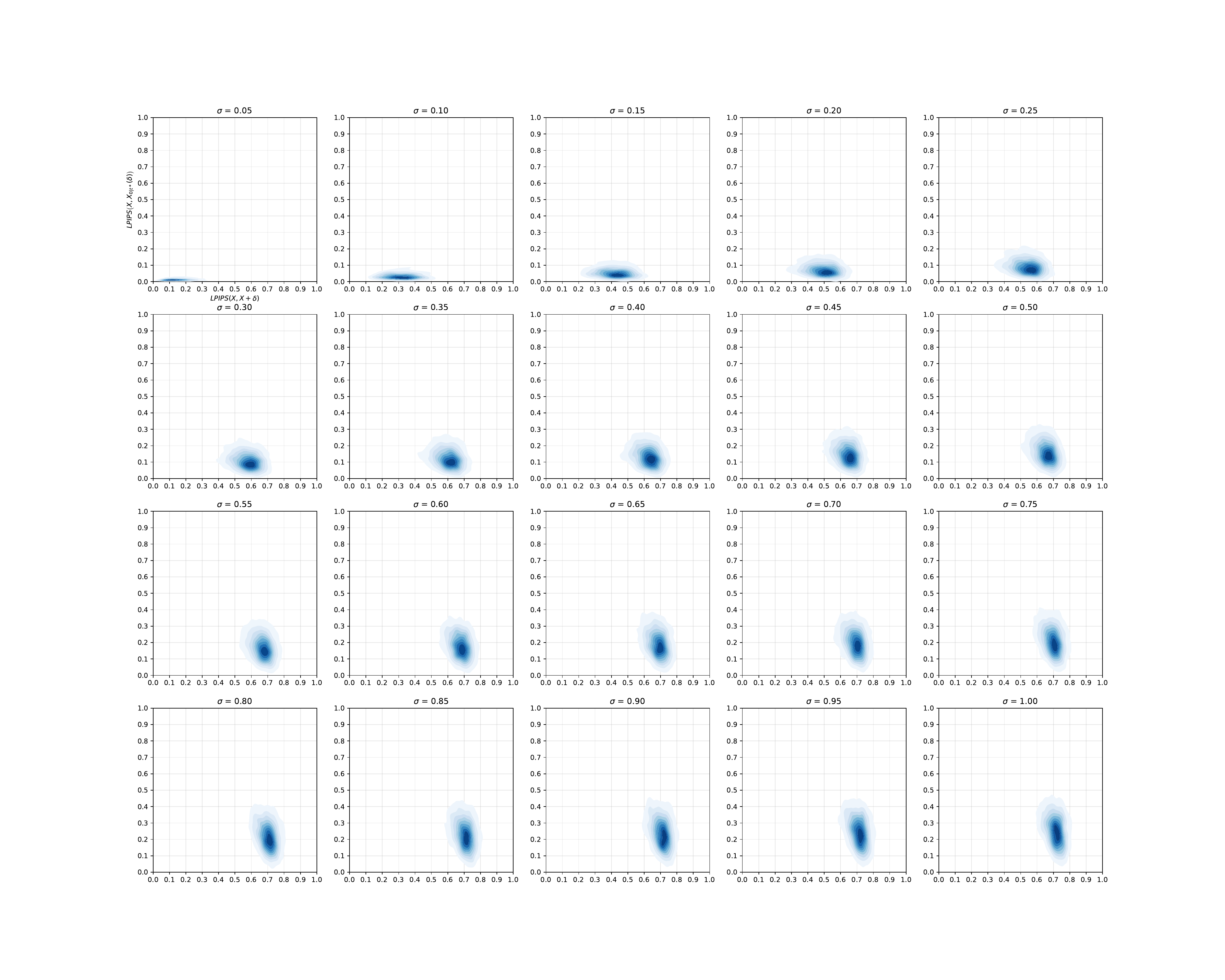}
	
    \caption{
        KDE plot of the LPIPS distance between clean data and the base classifier's input for randomized smoothing methods without a denoiser (x-axis) and the corresponding distance for diffusion-denoised smoothing (y-axis).
    }
    \label{cifar10_lpips_distances} 
\end{figure*}

To compare diffusion-denoised smoothing with other randomized smoothing methods that do not incorporate a denoiser before the base classifier, referred to as denoiserless methods, we evaluate the distance between the base classifier's input and the clean data. We utilize $l_2$ and LPIPS distances to compare these two datasets in terms of geometric and perceptual differences, respectively. In denoiserless methods, the $l_2$ distance between the base classifier's input and the clean data corresponds to $\|\delta\|_2$. For diffusion denoised smoothing, this distance for a given data point x is represented as $\|x - x_{0\lvert t}\left(\delta\right)\|$. Figures \ref{cifar10_l2_distances} and \ref{cifar10_lpips_distances} illustrate the relationship between these two distances for denoiserless methods and diffusion denoised smoothing as KDE plots, presented for different values of $\sigma$ for both $l_2$ and LPIPS distances, respectively.

Figure \ref{cifar10_l2_distances} demonstrates that using a denoiser before the base classifier maps the perturbed data closer to the clean data, a trend that is clearly observable. Consequently, in denoiserless methods, the absence of a denoiser can result in the base classifier's input becoming out-of-distribution, leading to poor performance by the base classifier. However, in diffusion-denoised smoothing, the denoising process produces data that is much closer to the clean data, which the base classifier has been trained on. Therefore, utilizing an effective denoiser, such as a denoising diffusion model, before the base classifier proves to be a beneficial choice.

Figure \ref{cifar10_l2_distances} also illustrates that the distance between the denoised data and the clean data remains significant in diffusion-denoised smoothing across all $\sigma$ values. This phenomenon is what we refer to as the covariate shift. Although the covariate shift is relatively small for lower $\sigma$ values, for higher $\sigma$ values, the single-shot denoiser introduces a significant covariate shift, which further increases as $\sigma$ grows. Therefore, it is essential to make the base classifier robust to covariate shifts, particularly for larger $\sigma$ values.

Figure \ref{cifar10_lpips_distances} presents the KDE plot of the LPIPS distance between the clean data and the base classifier's input for both denoiserless methods and diffusion-denoised smoothing. This figure shows that as $\sigma$ increases, the perceptual distance between the base classifier’s input and the clean data in denoiserless methods grows significantly faster compared to diffusion-denoised smoothing. This highlights the critical importance of incorporating a denoiser before the base classifier. We also observe that the perceptual distance between clean and noisy-then-denoised data in diffusion denoised smoothing becomes significant for large $\sigma$ values, which we previously identified as the covariate shift. Thus, it is crucial to make the base classifier robust to covariate shifts, particularly at large $\sigma$ values.

\section{Visual Interpretation of Extreme Covariate Shifted Data}
\begin{figure*}[!t]
    \centering 
    \includegraphics[width=0.9\textwidth, trim=0mm 50mm 0mm 50mm, clip]{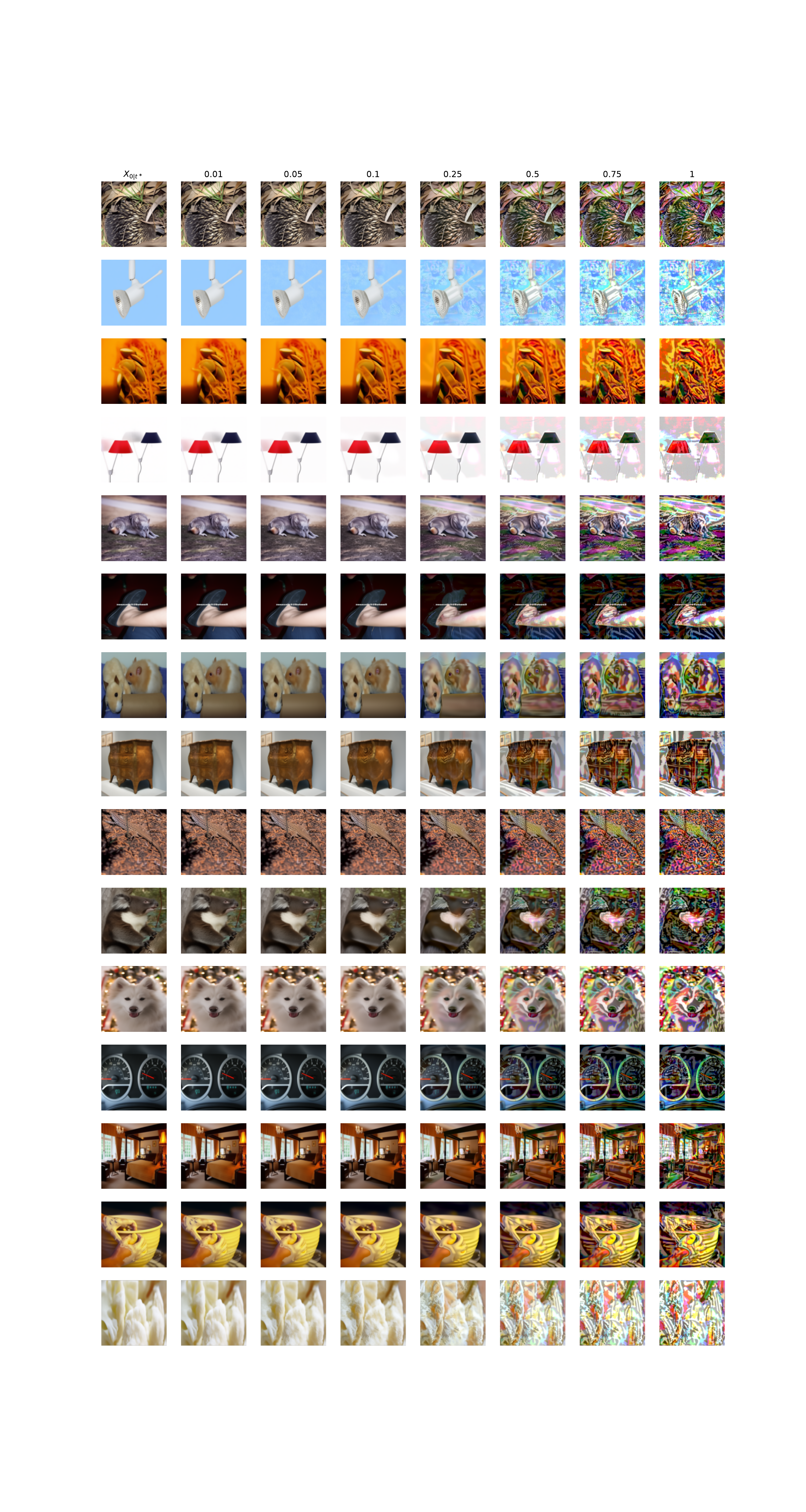}
	
    \caption{Outputs of Algorithm~\ref{extreme_covariate_shifted_data_finding_algorithm} with $M=1$, starting from the same $\epsilon_{0}$ but with different step sizes (indicated at the top). The leftmost column shows the noisy-then-denoised data obtained using $\epsilon_{0}$.
    }
    \label{imagenet_adv_1_step} 
\end{figure*}

\begin{figure*}[!t]
    \centering 
    \includegraphics[width=0.9\textwidth, trim=0mm 50mm 0mm 50mm, clip]{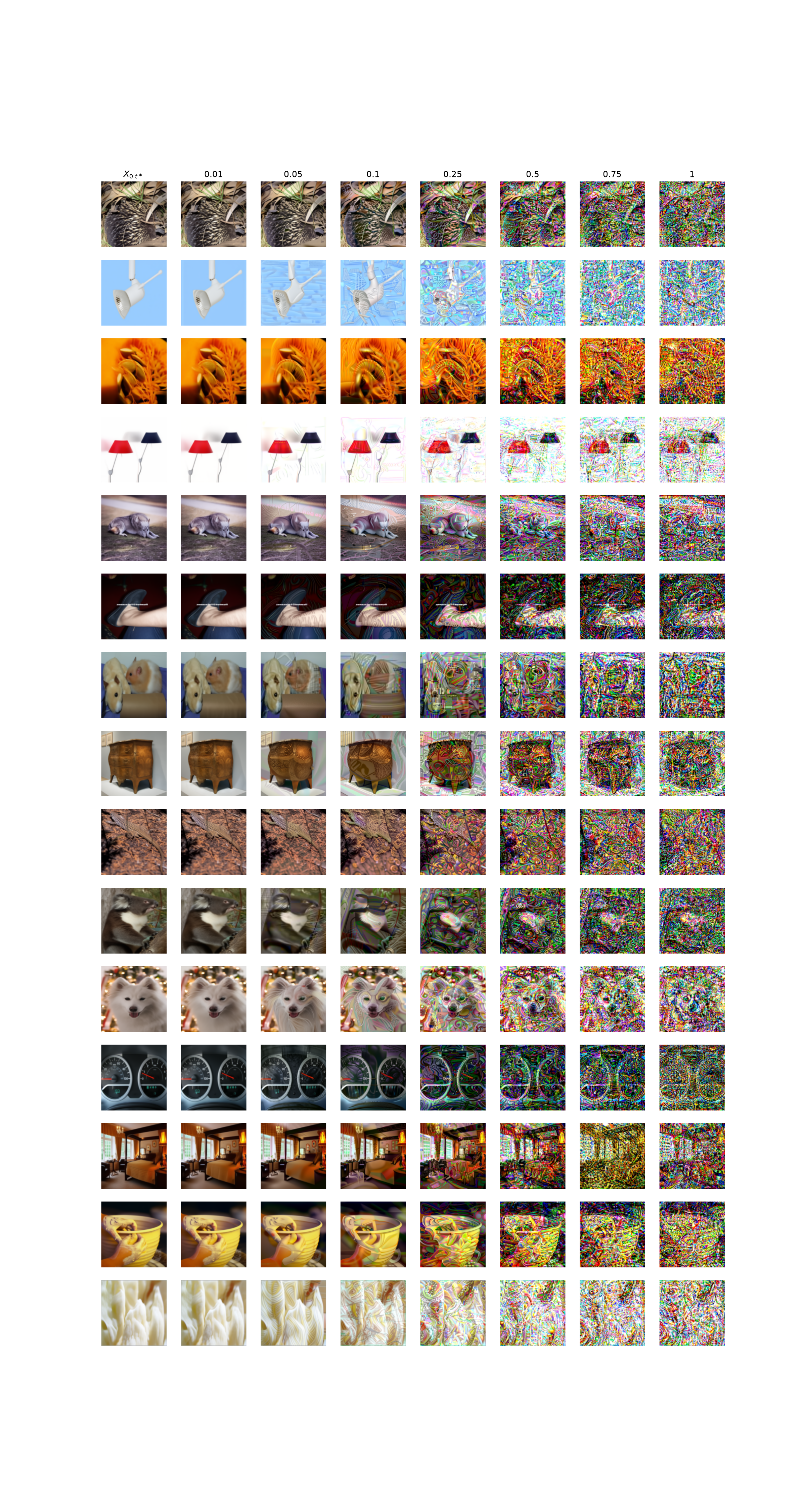}
	
    \caption{Outputs of Algorithm~\ref{extreme_covariate_shifted_data_finding_algorithm} with $M=10$, starting from the same $\epsilon_{0}$ but with different step sizes (indicated at the top). The leftmost column shows the noisy-then-denoised data obtained using $\epsilon_{0}$.
    }
    \label{imagenet_adv_10_steps} 
\end{figure*}

Figures \ref{imagenet_adv_1_step} and \ref{imagenet_adv_10_steps} illustrate the adversarial samples generated by Algorithm 1 for different values of the step size parameter and number of gradient ascent iterations $M$. The goal of the algorithm is to find visually-indistinguishable yet semantically misleading samples under the diffusion-denoising pipeline. Starting from a clean image $x$, we first add a random noise vector $\epsilon_0 \sim \mathcal{N}\left(0,I\right)$, apply a single-step denoising using the pretrained diffusion model $\tilde{\epsilon}_\theta$ and feed the resulting sample $x_{0\lvert t^\ast}$ into the classifier $f_{\phi}$. We then iteratively update the noise direction to maximize the classification loss, ensuring that the resulting sample remains within a bounded perturbation radius $r_{adv}$.  The final adversarially perturbed sample is obtained after $M$ steps.

In Figure \ref{imagenet_adv_1_step}, we show the effect of varying the step size $M=1$, while in Figure \ref{imagenet_adv_10_steps}, the same experiment is repeated for $M=10$. Each row corresponds to a different original image, and each column shows the final generated sample for a specific step size, ranging from 0.01 to 1.0. As seen in the figures, increasing the step size amplifies the degree of covariate shift in the diffusion-denoised space, eventually resulting in highly distorted samples that still manage to fool the classifier. The visual comparison across different values of $M$ also highlights the stronger adversarial effect achieved when the number of iterations is increased.

\begin{table*}[t]
    \centering
    \caption{Approximate certified test accuracy on MNIST}
    \label{approximate_certified_test_accuracy_table_mnist}
        \resizebox{\textwidth}{!}{
            \begin{tabular}{cl|ccccccccccc}
                \toprule
                {$\sigma$} & {Method (MNIST)} & {ACR}
                & 0.00 & 0.25 & 0.50 & 0.75 & 1.00 & 1.25 & 1.50 & 1.75 & 2.00 & 2.25 \\
                \midrule
                \multirow{12}{*}{0.25} & Randomized Smoothing& 0.716
                &96.6  &93.8&  85.6&  69.0&  0.0&  0.0&  0.0&  0.0&  0.0&  0.0
                \\
                &SmoothAdv& 0.745  & 98.6 &  96.2 &  92.0 &  79.6 &  0.0  &0.0  &0.0  &0.0  &0.0  &0.0 \\
                
                & MACER &0.918 & 99.0 & $\mathbf{99.0}$ & 97.0& 95.0 & 0.0 & 0.0 & 0.0 & 0.0 & 0.0 & 0.0 \\
        
                & SmoothMix & $\mathbf{0.933}$ & 99.4 & $\mathbf{99.0}$ & $\mathbf{98.2}$& $\mathbf{96.9}$ & 0.0 & 0.0 & 0.0 & 0.0 & 0.0 & 0.0 \\
                
                & Consistency & 0.928 & $\mathbf{99.5}$ & $\mathbf{98.9}$ & $\mathbf{98.0}$& 96.0 & 0.0 & 0.0 & 0.0 & 0.0 & 0.0 & 0.0 \\

                & Denoised Smoothing & 0.767 &    94.2&  92.4&  90.0&  85.4&  0.0&  0.0&  0.0&  0.0&  0.0  &0.0\\
                
                & DDS (Pretrained)& 0.784  &  94.0  & 92.8  & 92.0  & 90.2 &  0.0  & 0.0  & 0.0  & 0.0 &  0.0  & 0.0 \\
                & DDS (Finetuned)& 0.789 &  94.4 &  94.0 &  92.8 &  91.6 &  0.0 &  0.0 &  0.0 &  0.0 &  0.0 &  0.0 \\
                &Ours ($r_{adv}=+\infty$, $M=1$)&  0.780  &    94.6&  92.8&  91.6&  90.4&  0.0 &  0.0 &  0.0 &  0.0 &  0.0 &  0.0 \\
                &Ours ($r_{adv}=+0.1$, $M=1$)&  0.788  &    95.4&	94.8&	93.8&	91.4&	0.0&	0.0&	0.0&	0.0&	0.0&	0.0 \\
                &Ours ($r_{adv}=+\infty$, $M=2$)&  0.787  &    94.6&	93.8&	92.8&	91.4&	0.0&	0.0&	0.0&	0.0&	0.0&	0.0 \\
                &Ours ($r_{adv}=+\infty$, $M=4$)&  0.787  &    95.0&	94.4&	93.4&	91.4&	0.0&	0.0&	0.0&	0.0&	0.0&	0.0 \\
                \midrule
                \multirow{12}{*}{0.5} & Randomized Smoothing & 1.179& 97.4 & 94.6 & 90.2 & 82.6 & 70.6 & 50.6 & 23.8 & 0.0 & 0.0 & 0.0 \\
                & SmoothAdv & 1.245& 97.6 & 96.0 & 92.4 & 87.0 & 79.2 & 59.6 & 26.2 & 0.0 & 0.0 & 0.0 \\
                
                & MACER &1.583 & $\mathbf{99.0}$ & 98.0& 96.0 & 94.0 & 90.0 & 83.0 & 73.0 & 50.0 & 0.0 & 0.0\\
                
                & SmoothMix & $\mathbf{1.694}$ & 98.8 & 98.1 & 97.1& 95.3 & 92.7 & 88.3 & 81.7 & $\mathbf{69.5}$ & 0.0 & 0.0 \\
                
                & Consistency & 1.657 & $\mathbf{99.2}$ & $\mathbf{98.6}$ & $\mathbf{97.6}$ & $\mathbf{95.9}$ & $\mathbf{93.0}$ & 87.8 & 78.5 & 60.5 & 0.0 & 0.0 \\
                
                & Denoised Smoothing & 1.367 & 93.4 & 91.8 & 89.2 & 85.8 & 80.0 & 71.2 & 53.0 & 0.0 & 0.0 & 0.0 \\
                & DDS (Pretrained) & 1.542 & 94.2 & 93.2 & 92.4 & 91.0 & 90.4 & 88.8 & 86.6 & 0.0 & 0.0 & 0.0 \\
                & DDS (Finetuned) & 1.547 & 95.0 & 94.2 & 93.6 & 92.8 & 91.2 & 89.8 & 87.8 & 0.0 & 0.0 & 0.0 \\
                & Ours ($r_{adv}=+\infty$, $M=1$) & 1.546 & 96.0 & 95.4 & 94.6 & 93.4 & 92.8 & 90.4 & 87.2 & 0.0 & 0.0 & 0.0 \\
                & Ours ($r_{adv}=+0.1$, $M=1$) & 1.548 & 95.0&	94.0&	93.4&	92.6&	91.8&	90.2&	87.2&	0.0&	0.0&	0.0 \\
                & Ours ($r_{adv}=+\infty$, $M=2$) & 1.560 & 95.4&	94.6&	93.6&	93.2&	$\mathbf{93.0}$&	$\mathbf{92.0}$&	$\mathbf{90.2}$&	0.0&	0.0&	0.0 \\
                & Ours ($r_{adv}=+\infty$, $M=4$) & 1.554 & 95.6&	94.8&	94.2&	93.8&	$\mathbf{93.0}$&	91.0&	88.8&	0.0&	0.0&	0.0 \\
                \midrule
                
                \multirow{12}{*}{1.00} & Randomized Smoothing &1.094 & 77.0 & 69.6 & 62.4 & 52.2 & 42.8 & 32.6 & 18.6 & 10.0 & 05.6 & 03.2 \\
                & SmoothAdv &1.081 & 88.4 & 81.2 & 72.4 & 61.0 & 48.0 & 33.4 & 20.8 & 10.6 & 05.2 & 02.6 \\
                
                & MACER & 1.520 & 89.0 & 85.0 & 79.0& 75.0 & 69.0 & 61.0 & 54.0 & 45.0 & 36.0 & 28.0 \\
                & SmoothMix & 1.788 & $\mathbf{95.5}$ & 93.5 & 90.5 & 86.2 & 80.6 & 73.4 & 64.3 & 53.7 & 43.2 & 33.5 \\
                
                & Consistency &1.740 & 95.0 & 93.0 & 89.7 & 85.4 & 79.7 & 72.7 & 63.6 & 53.0 & 41.7 & 30.8 \\
                
                & Denoised Smoothing &0.715 & 88.0 & 84.0 & 77.2 & 66.6 & 0.0 & 0.0 & 0.0 & 0.0 & 0.0 & 0.0 \\
                & DDS (Pretrained) & 2.846 & 93.8 & 93.4 & 92.4 & 91.4 & 90.0 & 89.0 & 87.6 & 85.8 & 83.0 & 80.4 \\
                & DDS (Finetuned) &2.817 & 93.8 & 93.2 & 92.4 & 91.6 & 89.8 & 88.0 & 86.2 & 84.2 & 82.6 & 79.8 \\
                & Ours ($r_{adv}=+\infty$, $M=1$)  & 2.834 & 95.2 & $\mathbf{94.2}$ & $\mathbf{93.6}$ & $\mathbf{92.6}$ & 90.6 & 88.8 & 87.2 & 86.0 & 84.2 & 82.0 \\
                & Ours ($r_{adv}=+0.1$, $M=1$)  & 2.836 & 95.0&	93.8&	93.2&	92.4&	90.6&	89.4&	87.2&	85.6&	84.8&	82.0	 \\
            & Ours ($r_{adv}=+\infty$, $M=2$)  & 2.863 & 95.0&	$\mathbf{94.0}$&	93.2&	91.6&	$\mathbf{91.4}$&	$\mathbf{90.0}$&	$\mathbf{88.4}$&	$\mathbf{87.4}$&	$\mathbf{85.6}$&	$\mathbf{82.6}$	 \\
                
            & Ours ($r_{adv}=+\infty$, $M=4$)  & $\mathbf{2.886}$ & 93.8&	92.8&	92.4&	91.4&	90.6&	$\mathbf{90.2}$&	$\mathbf{88.4}$&	86.4&	84.4&	$\mathbf{82.4}$\\
                \bottomrule
        \end{tabular}
    }
\end{table*}

\clearpage
\begin{table*}[t]
    \centering
    \caption{Approximate certified test accuracy on CIFAR-10}
    \label{approximate_certified_test_accuracy_table_cifar10}
        \resizebox{\textwidth}{!}{
            \begin{tabular}{cl|ccccccccccc}
                \toprule
                {$\sigma$} & {Method (CIFAR-10)} & {ACR}
                & 0.00 & 0.25 & 0.50 & 0.75 & 1.00 & 1.25 & 1.50 & 1.75 & 2.00 & 2.25 \\
                \midrule
                \multirow{3}{*}{0.25} & DDS (Pretrained)&	0.524&	78.4&	61.4&	45.2&	26.4&	0.0&	0.0&	0.0&	0.0&	0.0&	0.0\\
                &DDS (Fine-Tuned)&	0.621&	$\mathbf{79.0}$&	$\mathbf{69.6}$&	57.0&	43.0&	0.0&	0.0&	0.0&	0.0&	0.0&	0.0\\
                
                &Ours&				$\mathbf{0.658}$&	76.4&	69.0&	$\mathbf{59.8}$&	$\mathbf{47.2}$&	0.0&	0.0&	0.0&	0.0&	0.0&	0.0\\
                \midrule
                \multirow{3}{*}{0.5} & DDS (Pretrained)&	0.804&	62.0&	51.6&	39.8&	30.6&	23.0&	15.8&	9.0&	0.0&	0.0&	0.0\\
                &DDS (Fine-Tuned)&	1.055&	72.8&	65.6&	55.8&	48.8&	42.8&	36.4&	26.8&	0.0&	0.0&	0.0\\
                &Ours&				$\mathbf{1.122}$&	$\mathbf{75.2}$&	$\mathbf{69.6}$&	$\mathbf{62.8}$&	$\mathbf{54.6}$&	$\mathbf{46.6}$&	$\mathbf{39.0}$&	$\mathbf{32.0}$&	0.0&	0.0&	0.0\\
                \midrule
                
                \multirow{3}{*}{1.00} & DDS (Pretrained)&	0.991&	38.6&	31.6&	26.8&	20.6&	15.6&	12.2&	8.8&	5.6&	3.6&	3.0\\
                &DDS (Fine-Tuned)&	1.578&	$\mathbf{61.6}$&	$\mathbf{56.0}$&	48.6&	43.6&	37.6&	32.8&	30.2&	26.8&	23.0&	19.0\\
                &Ours&				$\mathbf{1.774}$&	60.0&	54.0&	$\mathbf{49.0}$&	$\mathbf{45.2}$&	$\mathbf{42.2}$&	$\mathbf{39.2}$&	$\mathbf{36.0}$&	$\mathbf{32.0}$&	$\mathbf{27.2}$&	$\mathbf{22.2}$\\
                \bottomrule
        \end{tabular}
    }
\end{table*}
\begin{table*}[t]
    \centering
    \caption{Approximate certified test accuracy on ImageNet}
    \label{approximate_certified_test_accuracy_table_imagenet}
    \resizebox{\textwidth}{!}{
        \begin{tabular}{cl|cccccccc}
            \toprule
            {$\sigma$} & {Method (ImageNet)} & {ACR}
            & 0.0&	0.5&	1.0&	1.5&	2.0&	2.5&	3.0 \\
            \midrule
            \multirow{3}{*}{0.50} & DDS (Pretrained)		&1.250	&73	&62	&54	&40	&0	&0	&0\\
            &DDS (Fine-Tuned)		&1.311	&76	&68	&59	&49	&0	&0	&0\\
            &Ours				&$\mathbf{1.367}$	&$\mathbf{79}$	&$\mathbf{74}$	&$\mathbf{64}$	&$\mathbf{56}$	&0	&0	&0\\
            \midrule
            \multirow{3}{*}{1.00} & DDS (Pretrained)		&1.966	&59	&54	&47	&37	&30	&23	&13\\
            &DDS (Fine-Tuned)		&2.113	&67	&$\mathbf{62}$	&53	&47	&39	&32	&22\\
            &Ours				&$\mathbf{2.212}$	&$\mathbf{68}$	&60	&$\mathbf{57}$	&$\mathbf{49}$	&$\mathbf{40}$	&$\mathbf{38}$	&$\mathbf{29}$\\
            \bottomrule
        \end{tabular}
    }
\end{table*}

\end{document}